\documentclass[sigconf, authorversion]{acmart}
\usepackage{multirow}
\usepackage[utf8]{inputenc} 
\usepackage[T1]{fontenc}    
\usepackage{url}            
\usepackage{booktabs}       
\usepackage{amsfonts}       
\usepackage{nicefrac}       
\usepackage{xcolor}         
\usepackage{amsmath}
\usepackage{mathtools}
\usepackage{amsthm}
\usepackage{colortbl}
\usepackage{tablefootnote}
\usepackage{booktabs}
\usepackage{adjustbox}
\usepackage{wrapfig}
\usepackage{graphicx}     
\usepackage{pifont}  
\usepackage{hhline}
\usepackage{enumitem}
\usepackage{etoolbox}
\usepackage{listings}
\usepackage{microtype}
\usepackage{graphicx}
\usepackage{subfigure}
\usepackage{booktabs} 
\usepackage{multirow}
\usepackage{makecell}
\usepackage{hyperref}
\usepackage{url}
\usepackage[utf8]{inputenc} 
\usepackage[T1]{fontenc}    
\usepackage{url}            
\usepackage{booktabs}       
\usepackage{amsfonts}       
\usepackage{nicefrac}       
\usepackage{xcolor}
\usepackage{mathtools}
\usepackage{amsthm}
\usepackage{booktabs}
\usepackage{adjustbox}
\usepackage{wrapfig}
\usepackage{graphicx}     

\usepackage{xspace}
\usepackage{booktabs}
\usepackage{multirow}
\usepackage[table]{xcolor}
\usepackage{caption}

\usepackage{subcaption}   
\makeatletter  
\newcommand{\thickhline}{%
    \noalign {\ifnum 0=`}\fi \hrule height 1pt
    \futurelet \reserved@a \@xhline
}
\definecolor{mygray}{gray}{.9}
\definecolor{ggray}{RGB}{127,127,127}
\definecolor{reda}{RGB}{192,0,0}
\definecolor{redb}{RGB}{217,148,143}
\definecolor{myyellow}{RGB}{190,144,0}
\definecolor{mygreen}{RGB}{0,176,80}
\definecolor{myred}{RGB}{248,66,0}
\definecolor{myblue}{RGB}{30,90,100}
\definecolor{mygray1}{RGB}{245,245,245}

\definecolor{agent}{RGB}{218, 165, 32}
\definecolor{goods}{RGB}{220, 20, 60}
\definecolor{payment}{RGB}{34,139,34}
\definecolor{seller}{RGB}{70,130,180}
\definecolor{place}{RGB}{138,43,226}

\usepackage{xspace}
\usepackage{booktabs}
\usepackage{multirow}
\usepackage[table]{xcolor}
\usepackage{caption}
\usepackage{pifont}
\usepackage{subcaption} 
\usepackage{enumitem}
\AtBeginDocument{%
  }

\acmConference{}
\setcopyright{none}
\renewcommand{\footnotetextcopyrightpermission}[1]{}
\begin{document}

\title{Deepfake Detection Generalization 
with Diffusion Noise}

\author{Hongyuan Qi}
\email{qihy@zju.edu.cn}
\affiliation{%
  \institution{Zhejiang University}
  \city{Hangzhou}
  \state{Zhejiang}
  \country{China}
}

\author{Wenjin Hou}
\email{houwj17@gmail.com}
\affiliation{%
  \institution{Zhejiang University}
  \city{Hangzhou}
  \state{Zhejiang}
  \country{China}
}

\author{Hehe Fan}
\authornote{Corresponding author.}
\email{hehe.fan.cs@gmail.com}
\affiliation{%
  \institution{Zhejiang University}
  \city{Hangzhou}
  \state{Zhejiang}
  \country{China}
}

\author{Jun Xiao}
\email{Junx@zju.edu.cn}
\affiliation{%
  \institution{Zhejiang University}
  \city{Hangzhou}
  \state{Zhejiang}
  \country{China}
}

\renewcommand{\shortauthors}{Hongyuan Qi, Feifei Shao, Ming Li, Hehe Fan, Jun Xiao}
\begin{abstract}
Deepfake detectors face growing challenges in generalization as new image synthesis techniques emerge. 
In particular, deepfakes generated by diffusion models are highly photorealistic and often evade detectors trained on GAN-based forgeries. This paper addresses the generalization problem in deepfake detection by leveraging diffusion noise characteristics. 
We propose an Attention-guided Noise Learning (ANL) framework that integrates a pre-trained diffusion model into the deepfake detection pipeline to guide the learning of more robust features. Specifically, our method uses the diffusion model's denoising process to expose subtle artifacts: the detector is trained to predict the noise contained in an input image at a given diffusion step, forcing it to capture discrepancies between real and synthetic images, while an attention-guided mechanism derived from the predicted noise is introduced to encourage the model to focus on globally distributed discrepancies rather than local patterns. 
By harnessing the frozen diffusion model's learned distribution of natural images, the ANL method acts as a form of regularization, improving the detector's generalization to unseen forgery types. 
Extensive experiments demonstrate that ANL significantly outperforms existing methods on multiple benchmarks, achieving state-of-the-art accuracy in detecting diffusion-generated deepfakes. Notably, the proposed framework boosts generalization performance (\textit{e.g.}, improving ACC/AP by a substantial margin on unseen models) without introducing additional overhead during inference. Our results highlight that diffusion noise provides a powerful signal for generalizable deepfake detection. 
\end{abstract}



\maketitle

\section{Introduction}
\begin{figure}
        \centering
        \includegraphics[width=0.9\columnwidth]{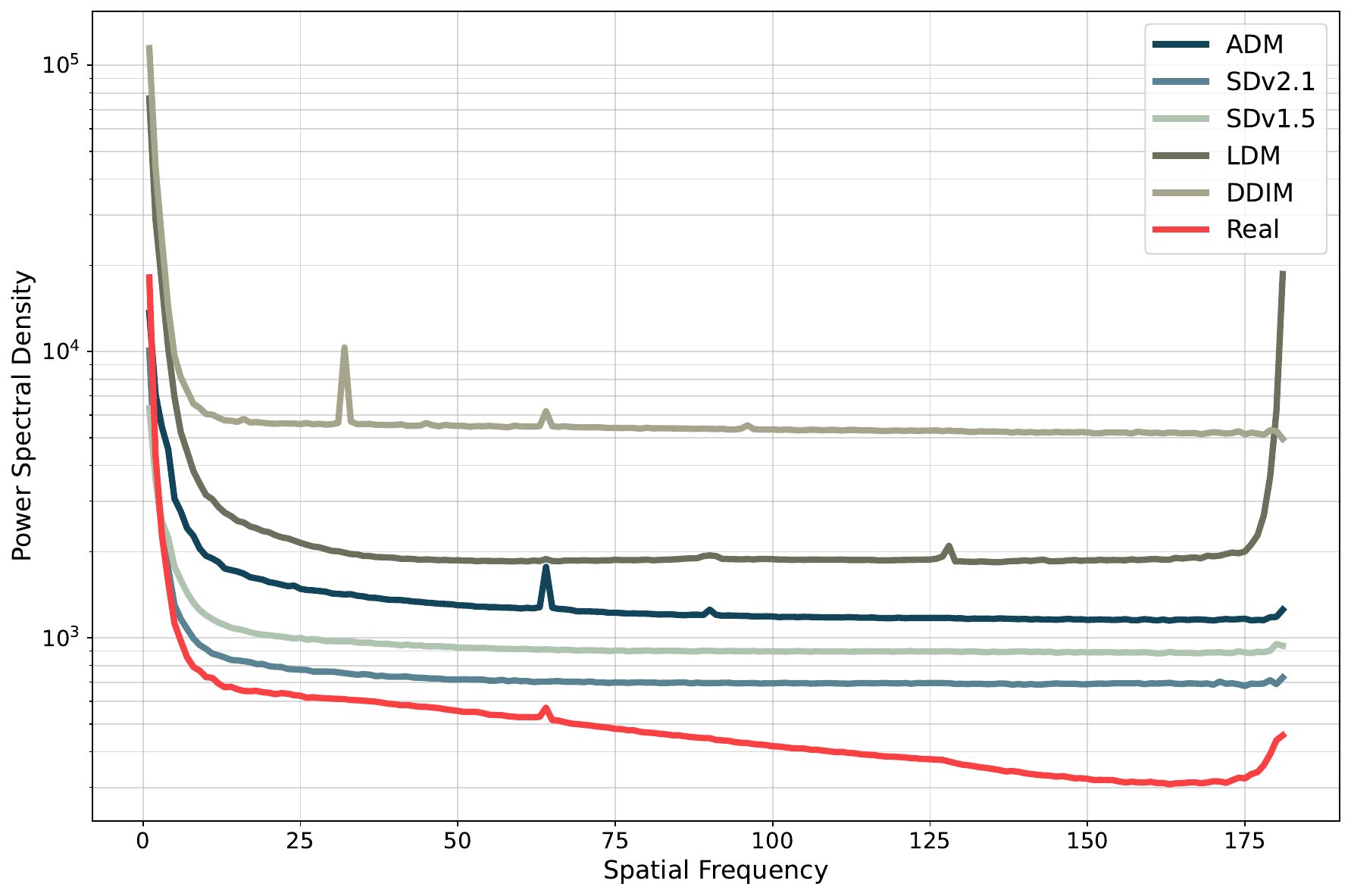}
        \caption{
            Power Spectral Density (PSD) analysis of noise domain features. Real images exhibit a noticeable drop in mid-frequency components, while diffusion-based fakes present a nearly flat distribution resembling white noise.
        }
        \label{fig:noise_visualization}
    \end{figure}
The rapid advancement of deepfake generation techniques has raised urgent concerns about security and misinformation~\cite{ho2020, dhariwal2021diffusionmodelsbeatgans}. 
Modern AI-generated content can produce highly realistic fake images and videos, making it increasingly difficult for humans – and even algorithms – to distinguish forgeries from authentic content~\cite{rossler2019faceforensicslearningdetectmanipulated, Cheng2024, chen2024, saharia2022photorealistictexttoimagediffusionmodels}. 
In particular, face swaps and other identity manipulations (commonly referred to as deepfakes) have become exceedingly convincing, fueling the spread of disinformation, privacy violations, and fraud~\cite{9712265, bateman2022deepfakes, westerlund2019emergence}. 
This escalation of deepfake realism necessitates robust detection methods that not only achieve high accuracy on known forgeries but also generalize to novel attacks encountered in the wild~\cite{liu2025evolvingsinglemodalmultimodalfacial, GouravGupta2024, DeepfakeBench_YAN_NEURIPS2023}. 
Unfortunately, many existing deepfake detectors struggle in generalization settings: a model trained on one type of forgery (\textit{e.g.}, a particular generative model or dataset) often fails to detect fakes from an unseen source, especially when the forgery technique or distribution differs from the training data. Addressing this challenge is now a critical focus in the research community, as it poses a significant barrier to their practical deployment in real-world scenarios.~\cite{ricker2024, song2023robustnessgeneralizabilitydeepfakedetection}. 

Early deepfake content was predominantly created using generative adversarial networks (GANs) or autoencoder-based models (VAEs), which often left behind telltale artifacts (such as frequency domain features or blending inconsistencies) that detectors could exploit~\cite{li2020facexraygeneralface, mccloskey2018detectinggangeneratedimageryusing, nataraj2019detectinggangeneratedfake, frank2020leveragingfrequencyanalysisdeep}.  
However, the landscape of generative models has evolved rapidly. Denoising diffusion models (and their latent variants such as Stable Diffusion) have emerged as a new paradigm for image synthesis, capable of producing extraordinarily realistic images with minimal obvious artifacts. Images forged by diffusion models tend to more closely follow the distribution of real images, blurring the line between real and fake~\cite{Wang2023, song2022, ma2023, Luo2024, Zhong2023}. 

Considering the unified iterative denoising process of diffusion models, a robust deepfake detector should satisfy two key requirements: (1) capture fundamental discrepancies between real and generated images during generation, enabling generalization across diverse and unseen models; and (2) reduce the influence of high-level semantics by focusing on subtle, low-level artifacts intrinsic to the diffusion process.

These fundamental and universal discrepancies in noise characteristics provide new insights for effective deepfake detection.
Motivated by these insights, we observe that real images, which do not undergo explicit denoising, inherently contain richer fine-grained details. When subjected to a single-step noise estimation using a diffusion model, these details are often treated as noise, producing structured and meaningful noise representations. In contrast, diffusion-generated images, produced through iterative denoising, yield noise estimates that resemble unstructured white noise (see Figure ~\ref{fig:noise_visualization}). These universal discrepancies in noise characteristics provide new insights for effective deepfake detection.

 In this work, we propose to harness the diffusion generative models that create advanced deepfakes as tools to detect them. 
 Our key insight is that pre-trained diffusion models carry rich knowledge of the natural image manifold, and an image that deviates from this manifold (as a forgery might) will induce telltale differences in the diffusion model's noise estimation process. 
 In other words, real and fake images can be distinguished by how a diffusion model ``perceives'' their noise. 
Building upon this critical observation, in this paper, we propose a novel \textbf{Attention-guided Noise Learning (ANL)} method particularly designed for detecting diffusion-generated deepfakes.
Instead of analyzing images directly, ANL harnesses a pre-trained diffusion model to predict the latent noise component within a given test image. 
By explicitly filtering out semantic information and guiding the model with noise-derived attention, ANL directs the detection model's attention to subtle, intrinsic generative artifacts embedded within noise patterns, thereby significantly enhancing robustness and generalization.
Additionally, we propose a cross-model evaluation protocol to show the generalization of our ANL against unseen generative models.
Extensive experiments validate that operating in the diffusion noise domain substantially improves generalization. In summary, our contributions are as follows:
\begin{itemize}[leftmargin=5mm]
\vspace{-1mm}
\item We propose a novel \textbf{Attention-guided Noise Learning (ANL)} method for detecting diffusion-generated deepfakes, which leverages diffusion-predicted noise to construct attention signals that guide the feature learning process.
\item We introduce a rigorous cross-model evaluation protocol targeting unseen generative methods, complementing standard and cross-dataset assessments to comprehensively evaluate generalization capability.
\item Extensive experiments across several benchmark datasets show that ANL significantly surpasses state-of-the-art methods in all evaluation settings. We further provide visual analyses of predicted noise patterns, revealing intrinsic distinctions between real and synthetic images.
\end{itemize}

\section{Related Work}

\subsection{Generative Diffusion Models}

The framework of diffusion models is inspired by concepts from nonequilibrium thermodynamics~\cite{sohldickstein2015deepunsupervisedlearningusing}, where data generation is conceptualized as a Markov chain process consisting of incremental noise addition (\textit{i.e.}, forward process) and gradual reconstruction of the original data through denoising (\textit{i.e.}, reverse process).
To address computational complexity and training challenges, Ho et al.~\cite{ho2020} propose Denoising Diffusion Probabilistic Models (DDPM), which simplify training by predicting the noise rather than modeling data distributions.
This approach notably improves training stability and image generation quality, establishing DDPM as a foundational framework extensively adopted in image synthesis tasks.

Subsequent efforts have focused on improving generation speed and fidelity. Denoising Diffusion Implicit Models (DDIM)~\cite{song2022} reinterpret diffusion as an ordinary differential equation (ODE) to enable faster sampling with fewer steps. 
Additionally, latent space optimization methods such as Stable Diffusion~\cite{rombach2022highresolutionimagesynthesislatent} employ VAEs to compress high-resolution images into a low-dimensional latent space, drastically reducing computational costs and enabling text-to-image synthesis. 
Multimodal extensions such as Imagen~\cite{saharia2022photorealistictexttoimagediffusionmodels} and DALL·E 2~\cite{ramesh2022hierarchicaltextconditionalimagegeneration} integrate cross-attention mechanisms, demonstrating robust conditional generation capabilities.  

Recently, diffusion models have expanded into dynamic and 3D content generation. 
Works like Make-A-Video~\cite{singer2022makeavideotexttovideogenerationtextvideo} and AnimatedDiff~\cite{guo2023animatediff} extend 2D frameworks to temporal domains for video synthesis. 
DreamFusion~\cite{poole2022dreamfusiontextto3dusing2d} leverages 2D priors for zero-shot 3D NeRF generation. 
As diffusion-based generative models continue to evolve, traditional deepfake detection methods struggle to remain effective, demanding new techniques tailored to the unique generative characteristics of diffusion.

\subsection{Diffusion Deepfake Detection}

Deepfake detection has achieved impressive progress, but most of the early research primarily focused on images generated by GANs or VAEs~\cite{frank2020leveragingfrequencyanalysisdeep, li2020facexraygeneralface, zhang2024learningnaturalconsistencyrepresentation, zhang2021detecting, yan2023ucfuncoveringcommonfeatures}. 
Recently, with the development of diffusion models, their surprising generative capabilities have motivated researchers to develop deepfake detection specifically targeting diffusion-generated images.
On one hand, existing detection models focus on specific artifacts inherent in their respective generation processes, leading to significant challenges when these methods are transferred to diffusion-generated images~\cite{ricker2024}. This highlights the necessity of detection methods specifically tailored to this new generative paradigm.
On the other hand, generalization has emerged as a critical issue due to the rapid increase of diffusion-based models~\cite{Lorenz2023, Cheng2024}, directly impacting the capability of detection models to handle unforeseen challenges in real-world settings.
\begin{figure*}[ht]
    \centering %
    \includegraphics[width=0.94\textwidth]{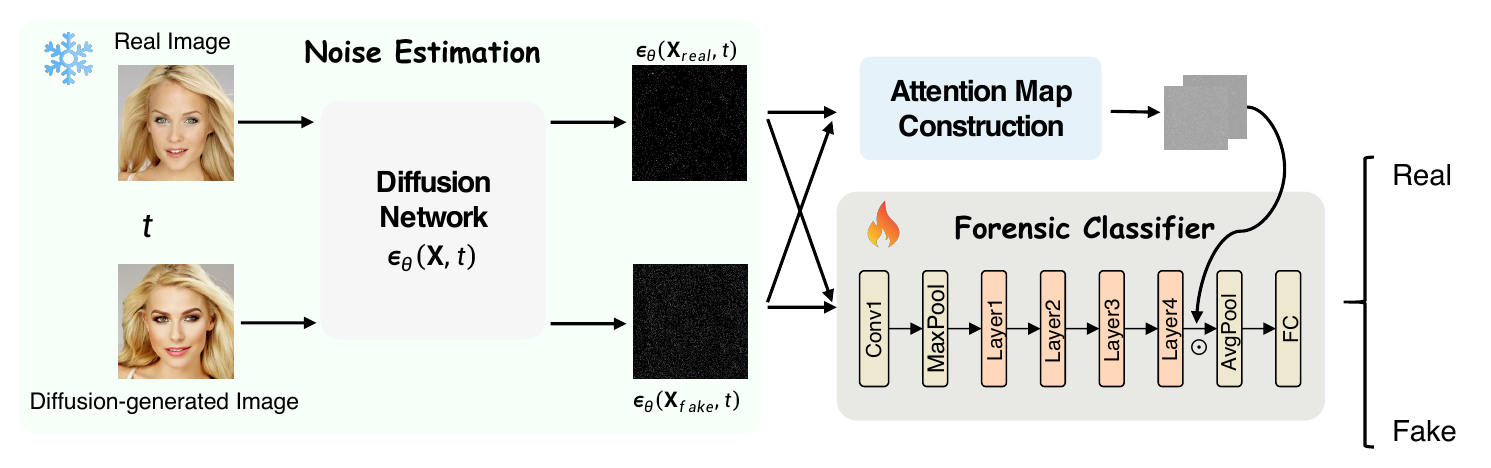} 
    \caption{Pipeline of Attention-guided Noise Learning (ANL). ANL implements noise estimation using the diffusion noise estimation network in the reverse process of DDPM~\cite{ho2020}. The predicted noise is then used to train a forensic classifier (\textit{e.g.}, ResNet \cite{he2015deepresiduallearningimage}) to distinguish between real and fake images, conbined with the attention map constrcted from the predicted noise. } 
    \label{fig:method} 
\end{figure*}
To address these issues, recent studies have proposed effective solutions from various perspectives. For example, DIRE~\cite{Wang2023} calculates the reconstruction error of images with DDIM~\cite{song2022}, and leverages this to achieve more generalized detection. SeDID~\cite{ma2023}, leveraging the SecMI method, computes the errors between reverse and denoised samples at a specific timestep in the generation process, integrating the features produced at intermediate steps of the diffusion model into the detection process. LaRE2~\cite{Luo2024} computes the reconstruction error in latent space to reduce the cost of feature extraction while preserving the essential information required for the detection of diffusion-generated images. Additionally, Zhong et al. ~\cite{Zhong2023} utilize the inter-pixel correlation contrast between rich and poor texture regions within an image as the universal fingerprint of synthetic images.

Despite their remarkable effectiveness, the above methods mainly address detection problems within the image domain. 
Moreover, regarding generalization, most of them only evaluate performance either within the same dataset or across different datasets (\textit{i.e.}, cross-dataset generalization). 
Evaluation on unseen models (cross-model generalization) remains largely unexplored. 
In this paper, we introduce a novel method that explicitly addresses this gap by operating in the noise domain. We demonstrate its strong generalization through extensive cross-model evaluations.

\section{Methodology}

In this section, we present our Attention-guided Noise Learning (ANL) method. As illustrated in Figure~\ref{fig:method}, ANL first leverages the reverse process of a DDPM to estimate the diffusion noise of a given image at a single timestep. Based on the predicted noise, we construct a spatial attention map that captures the distribution of noise intensity. This attention map is then used to guide the feature learning process of a classification network for distinguishing real images from diffusion-generated deepfakes.
To better introduce our method, we first briefly outline the fundamental principles of DDPM in Section~\ref{Preliminaries}. We then describe the noise estimation process in Section~\ref{np}, followed by the attention map construction in Section~\ref{subsec:attention} and its integration into the classifier in Section~\ref{subsec:classifier}. 
Note that we intentionally avoid complex data preprocessing or intricate model designs, aiming to clearly demonstrate the generalization capability of ANL.

\subsection{Preliminaries}
\label{Preliminaries}
Diffusion models synthesize data through a forward diffusion stage, in which noise is gradually introduced to training samples until they approximate a tractable prior distribution. 
Subsequently, these models learn a reverse diffusion process, iteratively denoising an initial random noise to generate novel samples that effectively capture the underlying characteristics of the original data distribution. 
We follow the standard notations and definitions in~\cite{ho2020}. 
Formally, diffusion models involve a $T$-step denoise process to generate images approximating the real data distribution. 
The forward diffusion process is defined step-by-step as:
\begin{equation}
    \label{eq1}
        q(\mathbf{x}_t|\mathbf{x}_{t-1})=\mathcal{N}(\mathbf{x}_t;\sqrt{1-\beta_t}\mathbf{x}_{t-1},\beta_t\mathbf{I}),
\end{equation}
where $\mathbf{x}_t$ denotes the noisy image at step $t$ generated from $\mathbf{x}_{t-1}$, and $\beta_t$ is a predefined noise scale parameter for each timestep $t\in \left [ 1,T \right ] $. 
By applying the Markov property of the forward process, we obtain:
\begin{equation}
    \label{eq3}
    q(\mathbf{x}_t|\mathbf{x}_0)=\mathcal{N}(\mathbf{x}_t;\sqrt{\bar{\alpha}_t}\mathbf{x}_0,(1-\bar{\alpha}_t)\mathbf{I}),
\end{equation}
where $\mathbf{x}_t$ refers to the diffusion product at step $t$, ${\alpha}_t=1-{\beta}_t$ and
$\bar{\alpha}_t=\prod_{s=1}^t\alpha_s$.

The reverse diffusion process aims to iteratively remove noise, defined by:
\begin{equation}
    \label{eq2}
    p_\theta(\mathbf{x}_{t-1}|\mathbf{x}_t)=\mathcal{N}(\mathbf{x}_{t-1};\boldsymbol{\mu}_\theta(\mathbf{x}_t,t),\boldsymbol{\Sigma}_\theta(\mathbf{x}_t,t)).
\end{equation}
where $\theta$ represents model parameters. 
The mean and variance parameters are defined as follows:
\begin{equation}
    \label{eq4}
    \boldsymbol{\Sigma}_\theta(\mathbf{x}_t,t)=\sigma_{t}^{2}\mathbf{I}, \sigma_{t}^{2}=\frac{1-\bar{\alpha}_{t-1}}{1-\bar{\alpha}_{t}}\beta_{t},
\end{equation} 
\begin{equation}
    \label{eq5}
    \boldsymbol{\mu}_\theta(\mathbf{x}_t,t)=\frac{1}{\sqrt{\alpha_t}}\left(\mathbf{x}_t-\frac{\beta_t}{\sqrt{1-\bar{\alpha}_t}}\boldsymbol{\epsilon}_\theta(\mathbf{x}_t,t)\right),
\end{equation} 
where $\boldsymbol{\epsilon}_\theta(\mathbf{x}_t,t)$ is a neural network-based function approximator trained to predict the noise component from the noise input $\mathbf{x}_t$ at timestep $t$.  
Diffusion models generate novel samples by iteratively executing this reverse diffusion process from timestep $T$ down to $1$.
In our method, we utilize a pre-trained $\boldsymbol{\epsilon}_\theta(\mathbf{x}_t,t)$ network to produce noise outputs of target images.
\begin{figure*}[h] 
    \centering 
    \vspace{-4mm}
    \includegraphics[width=0.9\textwidth]{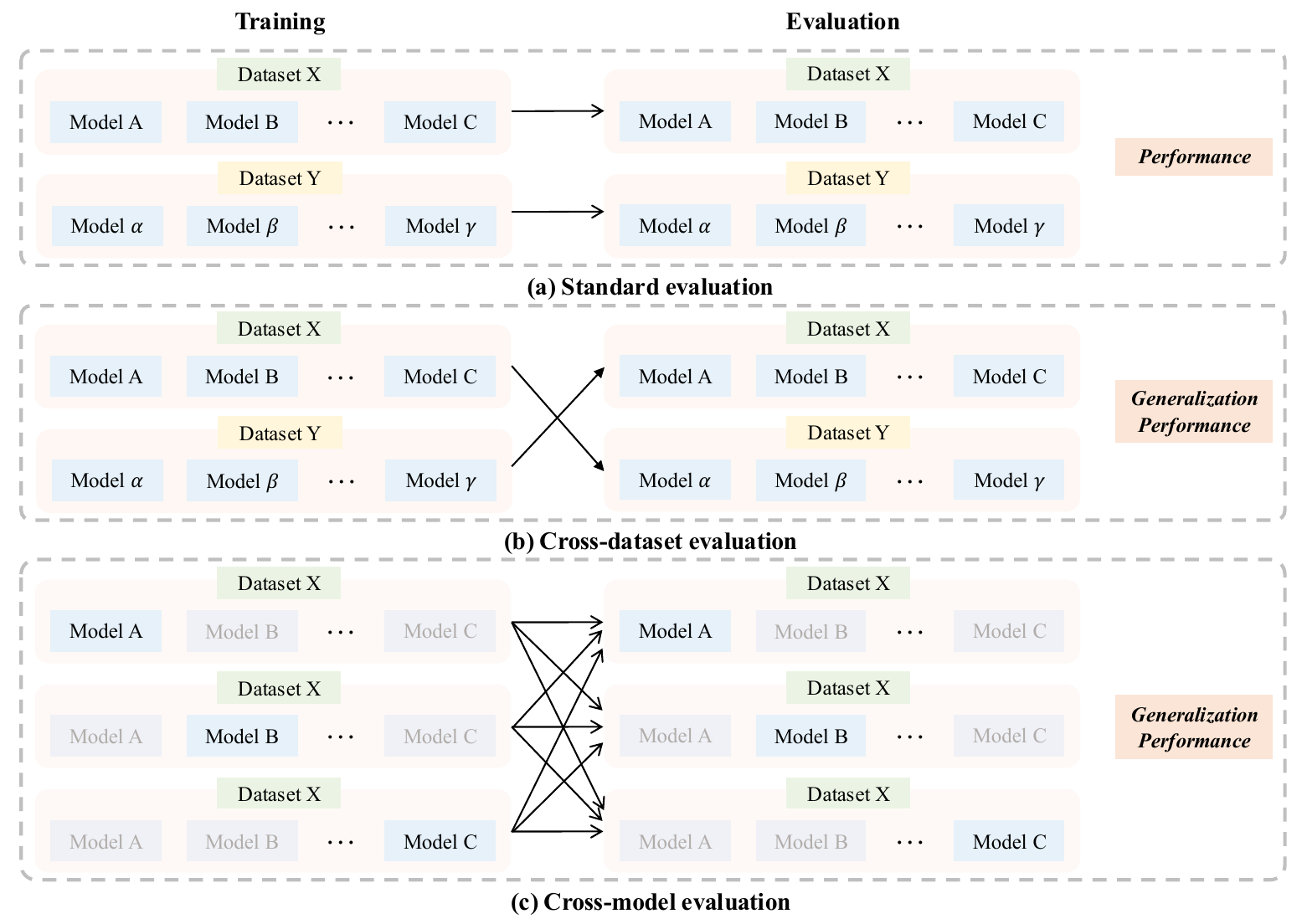} 
    \caption{Evaluation comparison under standard, cross-dataset, and cross-model scenarios.} 
    \label{fig:evaluation} 
    \vspace{-3mm}
\end{figure*}
\subsection{Noise Estimation}
\label{np}
In this paper, we identify a crucial distinction between real and diffusion-generated fake images in the diffusion noise domain:
\begin{itemize}[leftmargin=5mm]
    \item \textit{Real images} have not undergone any explicit denoising procedure, thus retaining inherently richer fine-grained details. 
    When subjected to a single-step noise estimation using a diffusion model, these detailed textures tend to be identified as noise by the model, resulting in predicted noise containing meaningful latent patterns.
    
    \item \textit{Diffusion-generated deepfakes}, in contrast, result from extensive iterative denoising processes. Consequently, when an additional noise estimation step is applied, these images contain fewer elements recognized as noise, leading to predicted noise that closely resembles white noise.
\end{itemize}
Based on this key observation, we propose our Attention-guided Noise Learning (ANL) method to effectively detect diffusion-generated deepfakes.
As illustrated in Figure ~\ref{fig:method}, we harness the noise estimation network $\boldsymbol{\epsilon}_\theta$ from the DDPM framework defined in Eq. (\ref{eq5}) to predict the noise component present in both real and diffusion-generated images. 
Given an input image, we treat it as a noise sample $\mathbf{x}$ at a specified timestep $t$. 
The noise predicted from DDPM's noise estimation network $\boldsymbol{\epsilon}_\theta$ can be represented as:
\begin{equation}
    \boldsymbol{\epsilon}_\theta(\mathbf{x},t),
    \label{eq:noise_prediction}
\end{equation}
where $t\in \left [ 1,T \right ] $. Different from the complete image generation process typical of DDPMs, our method does not execute the full reverse diffusion process.
We operate exclusively at a single timestep $t$, independent of the complete reverse diffusion procedure.
Furthermore, rather than aiming to produce a denoised image, our method directly outputs the noise component that the diffusion model estimates is present within the input image at this specified timestep. We detail the strategy selection for timestep $t$ in the Figure~\ref{fig:timestep}. 

\subsection{Attention Map Construction}
\label{subsec:attention}
After obtaining these estimated noise representations $\boldsymbol{\epsilon}_\theta(\mathbf{x},t)$ for both real images and diffusion-generated deepfakes, a straightforward approach is to directly train a binary forensics classifier (\textit{e.g.}, ResNet \cite{he2015deepresiduallearningimage}) to distinguish between the two classes based on their noise patterns. However, directly feeding the predicted noise into a standard CNN does not explicitly emphasize its global spatial distribution. As convolutional networks primarily rely on local receptive fields~\cite{he2015deepresiduallearningimage}, noise patterns that provide global cues for forensics may be more difficult to capture than local variations. As a result, the model may rely more on local patterns during training.

To address this issue, instead of using the estimated noise as a direct input, we further exploit its spatial structure to guide feature learning in the classification network. The predicted noise is thus used as an intermediate representation, from which a spatial attention signal is derived to modulate the feature responses of the backbone network.
\begin{figure*}[!t]
  \centering
  \begin{minipage}[t]{0.495\textwidth}
    \centering
    \includegraphics[width=\textwidth]{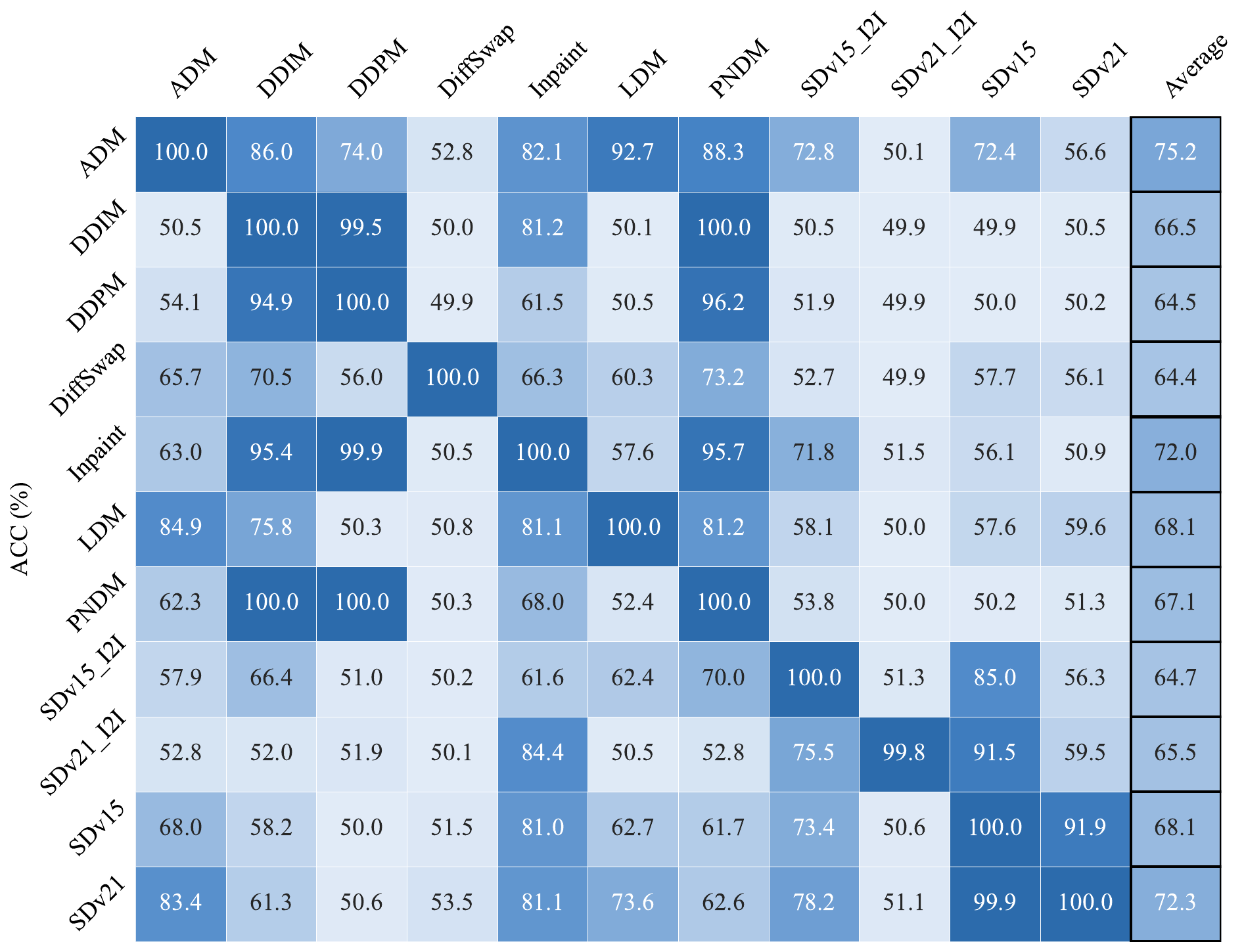}
    \smallskip
    \textbf{(a)} DIRE
  \end{minipage}
  \hfill
  \begin{minipage}[t]{0.495\textwidth}
    \centering
    \includegraphics[width=\textwidth]{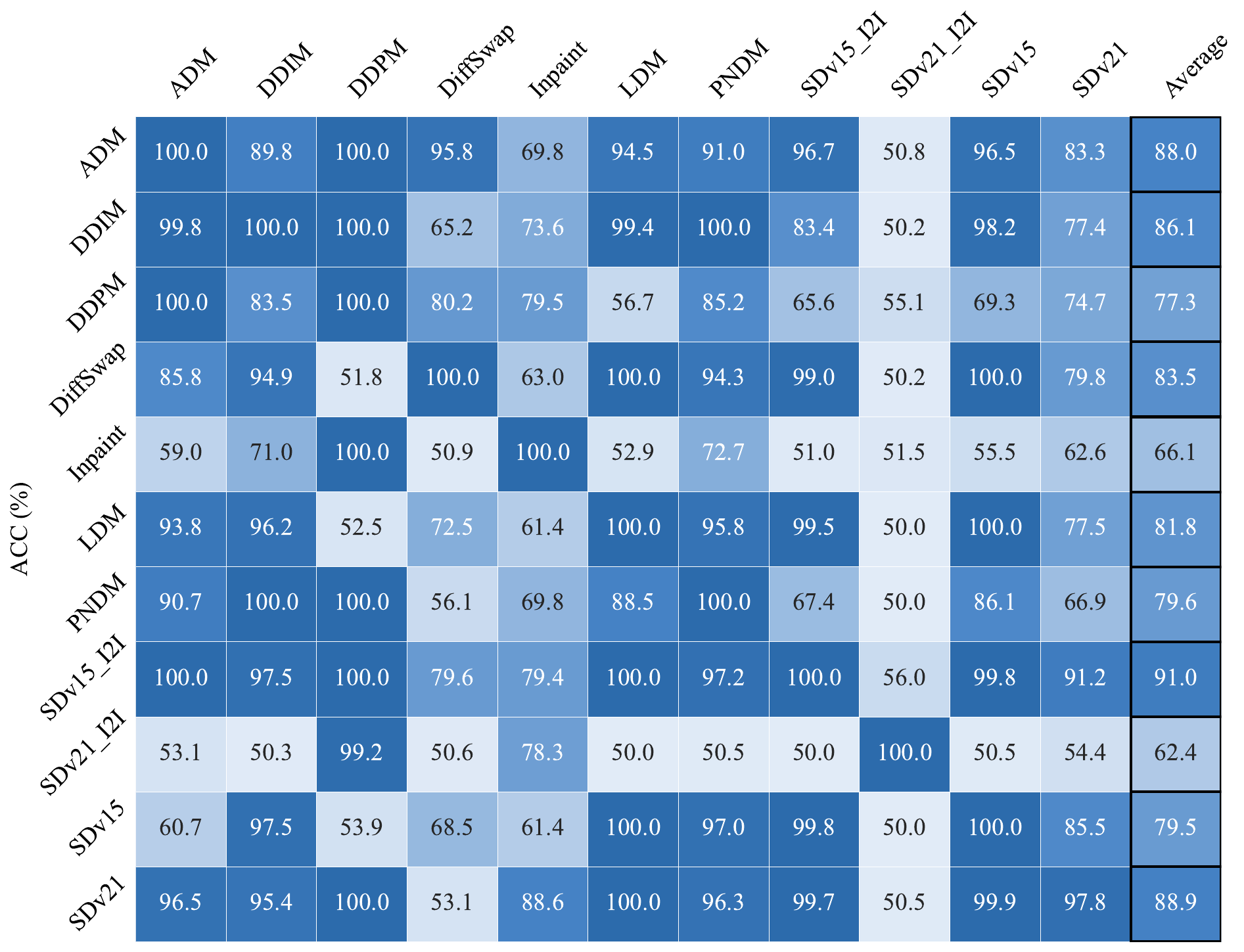}
    \smallskip
    \textbf{(b)} ANL
  \end{minipage}

  \caption{Cross-model evaluation results (ACC) on DiffFace. Labels on the left indicate the generative models used for \textbf{training} (\textit{i.e.}, data generated by these models), while the labels at the top represent the generative models used for \textbf{testing}. Darker colors indicate better performance. Clearly, our method exhibits significantly stronger generalization than the state-of-the-art DIRE \cite{Wang2023} under the cross-model scenario. More results of ACC and AP are provided in Appendix.
}
  \label{fig:cross_model_heatmap}
  \vspace{-3mm}
\end{figure*}
We construct a spatial attention map from the predicted noise. Specifically, we first compute the absolute value of the predicted noise and aggregate it across channels to obtain a spatial map:
\begin{equation}
A(\mathbf{x}) = \mathcal{N}\left( \frac{1}{C} \sum_{c=1}^{C} \left| \boldsymbol{\epsilon}_\theta(\mathbf{x},t)_c \right| \right),
\end{equation}
where $\mathcal{N}(\cdot)$ denotes a normalization operation that scales the values into $[0,1]$, and $C$ is the number of channels. The resulting $A(\mathbf{x}) \in [0,1]^{H \times W}$ reflects the spatial distribution of noise intensity.

\subsection{Attention-guided Classifier}
\label{subsec:classifier}

After obtaining the attention map $A(\mathbf{x})$, we use it to guide the feature learning process of the classification network. We adopt a standard ResNet backbone \cite{he2015deepresiduallearningimage} as the binary classifier.

Given an input image, the backbone extracts hierarchical feature representations and take the feature map $\mathbf{F}$ from the last stage of the network. To align with the feature resolution, we resize $A(\mathbf{x})$ and apply it to the feature map via element-wise multiplication:
\begin{equation}
\mathbf{F}' = \mathbf{F} \odot A(\mathbf{x}),
\end{equation}
where $\odot$ denotes element-wise multiplication.

The reweighted feature $\mathbf{F}'$ is then fed into the global pooling and classification layers to produce the final prediction. In this way, the classifier is guided to focus on regions with stronger noise responses, while suppressing less informative areas, leading to more robust feature representations.

To keep the classification model simple, we modify only the final layer of the ResNet, replacing it with a linear layer to produce a binary classification output.
\subsection{Loss Function and Inference}
\paragraph{Loss Function.}
Let $N$ represent the number of predicted noise samples in a given batch.
For each noise sample $i$ ($i=1, \dots, N$), let $y_i \in \{0, 1\}$ denote the ground truth label indicating fake (1) or real (0). The predicted probability $p_i$ of sample $i$ belonging to the fake class is obtained via sigmoid activation: $p_i = \sigma(z_i) = \frac{1}{1 + e^{-z_i}}$, where $z_i$ is the raw model output for sample $i$.
The Binary Cross-Entropy loss, denoted as $L_{\text{BCE}}$, is then calculated as follows:

\begin{equation}
L_{\text{BCE}} = - \frac{1}{N} \sum_{i=1}^{N} \left[ y_i \log(p_i) + (1 - y_i) \log(1 - p_i) \right].
\label{eq:bce_loss}
\end{equation}

\paragraph{Inference.}
After training the classifier, we perform inference as follows. 
Specifically, the target image $\mathbf{x}$ is first resized to $256\times256$. 
The noise representation is then predicted using Eq.~(\ref{eq:noise_prediction}) at a specified timestep $t$. 
Subsequently, this predicted noise is fed into the forensic classifier to obtain the final classification result, indicating whether the test image is real or fake.

\section{Experiments}
In this section, we provide a comprehensive evaluation to show the generalization of our ANL.
Current evaluation methods primarily focus on generalization within the same dataset or across different datasets. 
In this paper, we further propose cross-model generalization evaluation to better simulate real-world scenarios, especially given the rapid emergence of various diffusion-based generative models. 
Therefore, in Section~\ref{sec:comparison}, we first compare these evaluation scenarios.
Section~\ref{sec:setup} introduces our experimental setup.
In Section~\ref{sec:cross_model}, Section~\ref{sec:cross_dataset}, and Section~\ref{sec:standard}, we conduct experiments on cross-model evaluation, cross-dataset evaluation, and standard evaluation, respectively.
In Section~\ref{sec:ablation}, we perform ablation studies on the key components of our method.
Finally, in Section~\ref{sec:visualization}, we provide qualitative visualizations to further illustrate the effectiveness of our proposed approach.
\begin{table*}[!t]
\centering
\begin{minipage}{0.48\textwidth}
    \centering \setlength{\tabcolsep}{9pt}\small
   \caption{Cross-model average results. The best and second-best results are marked in bold and underline.}\label{tab:cross-model_performance}
\resizebox{1\textwidth}{!}{
\begin{tabular}{|r||cc|cc|}
\hline\thickhline
\rowcolor{mygray} & \multicolumn{2}{c|}{DiffFace} & \multicolumn{2}{c|}{DiFF} \\
\rowcolor{mygray}\multirow{-2}[-1]{*}{Method}
                        & ACC             & AP               & ACC             & \multicolumn{1}{c|}{AP}             \\
\hline\hline
ResNet18~\cite{he2015deepresiduallearningimage} & 60.06         & 68.00         & 65.20         & 70.54         \\
ResNet50~\cite{he2015deepresiduallearningimage} & 62.20         & 69.36        & 66.53         & 71.09         \\
ResNet101~\cite{he2015deepresiduallearningimage} & 64.42         & 73.06         & 67.12         & 74.64         \\
Xception~\cite{chollet2017xceptiondeeplearningdepthwise} & 65.90         & 72.78         & 65.79         & 71.77         \\
F3Net~\cite{qian2020thinkingfrequencyfaceforgery} & 65.54         & 66.85         & 67.29         & 77.16         \\
SeDID~\cite{ma2023} & 60.83         & 64.64         & 64.22         & 68.39        \\
NPR~\cite{tan2023rethinkingupsamplingoperationscnnbased} & 66.24         & 83.81         & \underline{68.02}         & 80.87         \\
DIRE~\cite{Wang2023} & \underline{68.05}         & \underline{88.74}        & 67.17         & \underline{82.71}         \\
\textbf{ANL (Ours)}          & \textbf{80.38}         & \textbf{95.61}        & \textbf{72.53}         & \textbf{92.52}         \\
\hline
\end{tabular}
} 
\end{minipage}\hfill
\begin{minipage}{0.48\textwidth}
    \centering \setlength{\tabcolsep}{9pt}\small
    \caption{Cross-dataset performance. The best and second-best results are marked in bold and underline.}\label{tab:cross-dataset_performance}
\resizebox{1\textwidth}{!}{
\begin{tabular}{|r||cc|cc|}
\hline\thickhline
\rowcolor{mygray}& \multicolumn{2}{c|}{DiffFace $\rightarrow$ DiFF} & \multicolumn{2}{c|}{DiFF $\rightarrow$ DiffFace} \\
\rowcolor{mygray}\multirow{-2}[-1]{*}{Method}  & ACC             & AP               & ACC             & \multicolumn{1}{c|}{AP}              \\
\hline\hline
ResNet18~\cite{he2015deepresiduallearningimage}  & 60.97         & 62.22   & 64.30         & 67.43              \\
ResNet50~\cite{he2015deepresiduallearningimage}   & 61.77         & 71.17   & 65.32         & 71.94               \\
ResNet101~\cite{he2015deepresiduallearningimage} & 60.85         & 72.92    & 65.71         & 77.58                \\
Xception~\cite{chollet2017xceptiondeeplearningdepthwise}  & 64.23         & 70.19    & 68.71         & 80.01             \\
F3Net~\cite{qian2020thinkingfrequencyfaceforgery} & 67.33         & 73.09       & 75.12         & 72.19           \\
SeDID~\cite{ma2023}   & 62.86         & 77.63        & 66.24         & 76.88        \\
NPR~\cite{tan2023rethinkingupsamplingoperationscnnbased} & 67.21         & 82.00       & 68.64         & 88.75           \\
DIRE~\cite{Wang2023}  & \underline{69.85}         & \underline{89.17}        & \underline{78.94}         & \underline{89.76}         \\
\textbf{ANL (Ours)}           & \textbf{82.61}         & \textbf{93.27}       & \textbf{92.96}         & \textbf{96.22}          \\
\hline
\end{tabular}
} 
\end{minipage}

\end{table*}

\begin{table*}[htbp]
  \centering
  \centering \setlength{\tabcolsep}{18pt}\small
  \caption{Standard evaluation results. The best and second-best results are in bold and underline.}\label{standard}
  \label{tab:performance}
  \begin{tabular}{|r||cc|cc|cc|}
    \hline\thickhline
     \rowcolor{mygray}& \multicolumn{2}{c|}{DiffFace} & \multicolumn{2}{c|}{DiFF} & \multicolumn{2}{c|}{DiffusionForensics} \\
    
     \rowcolor{mygray}\multirow{-2}[-1]{*}{Method}& ACC & \multicolumn{1}{c|}{AP} & ACC & \multicolumn{1}{c|}{AP} & ACC & \multicolumn{1}{c|}{AP} \\
    \hline\hline
    ResNet18~\cite{he2015deepresiduallearningimage} & 98.21 & 99.90 & 94.96 & 98.01 &  98.48 & \textbf{100.00} \\
    ResNet50~\cite{he2015deepresiduallearningimage} & 98.99 & \underline{99.99} & 98.92 & \textbf{100.00} & 98.12 & \textbf{100.00} \\
    ResNet101~\cite{he2015deepresiduallearningimage} & 98.51 & 99.98 & 98.79 & \textbf{100.00} & 98.89 & \textbf{100.00} \\
    Xception~\cite{chollet2017xceptiondeeplearningdepthwise} & 98.89 & 99.98 & 98.87 & \underline{99.99} & 98.82 & \textbf{100.00} \\
    F3Net~\cite{qian2020thinkingfrequencyfaceforgery} & 98.99 & \textbf{100.00} & 98.47 & \textbf{100.00} &  98.34 & \textbf{100.00} \\
    SeDID~\cite{ma2023} & 99.27 & \textbf{100.00} & 98.89 & \textbf{100.00} & 96.35 & \underline{99.96} \\
    NPR~\cite{tan2023rethinkingupsamplingoperationscnnbased} & \textbf{99.64} & \textbf{100.00} & 97.26 & \textbf{100.00} &  99.17 &\textbf{100.00}\\
    DIRE~\cite{Wang2023} & 98.80 & \textbf{100.00} & \underline{99.02} & \underline{99.99} &  \textbf{100.00}&\textbf{100.00}\\
    \textbf{ANL (Ours)} & \underline{99.55} & \textbf{100.00} & \textbf{99.52} & \textbf{100.00} &  \underline{99.48} & \textbf{100.00} \\
    \hline
  \end{tabular}
\end{table*}

\subsection{Comparison of Evaluation Methods}
\label{sec:comparison}
\paragraph{Standard Evaluation.} 
As shown in Figure \ref{fig:evaluation} (a), standard deepfake detection evaluation involves training and testing on the same dataset. Despite achieving impressive accuracy in this evaluation setting, real-world deepfakes may originate from generative models that are not represented in training dataset or are entirely unknown.

\paragraph{Cross-dataset Evaluation.} 
Cross-dataset evaluation involves training a detection model on dataset X and evaluating its performance on dataset Y, as illustrated in Figure \ref{fig:evaluation} (b). Recent studies have adopted this evaluation strategy to demonstrate the generalization capabilities of their proposed methods~\cite{Wang2023, tan2023rethinkingupsamplingoperationscnnbased, Luo2024}. However, the generative models present in these datasets may overlap. Thus, this evaluation does not sufficiently reflect the generalization of the detector, particularly against unseen generative models.

\paragraph{Cross-model Evaluation.} 
To address these limitations, we propose a cross-model evaluation strategy, as illustrated in Figure~\ref{fig:evaluation} (c). 
Specifically, the detection model is trained on images generated by a certain set of diffusion-based generative models (\textit{e.g.}, model A) and then evaluated on images generated by entirely different, unseen generative models (\textit{e.g.}, model B). 
This strategy serves as a rigorous test to validate the generalization of detection models, effectively capturing their ability to handle real-world scenarios involving unseen generative models.
We adopt this evaluation to assess the generalization of our proposed ANL method.

\subsection{Experimental Setup}
\label{sec:setup}
\paragraph{Diffusion Facial Datasets.}
Considering the higher real-world risks associated with facial deepfakes, we evaluate our method using two face datasets, \textbf{DiffFace} \cite{chen2024} and \textbf{DiFF} \cite{Cheng2024}, comprising images from multiple advanced diffusion models.
DiffFace~\cite{chen2024} covers various forgery classes (unconditional, text-guided, Img2Img, inpainting, and face-swapping) generated by 11 advanced diffusion models, including DDPM~\cite{ho2020}, ADM~\cite{dhariwal2021diffusionmodelsbeatgans},  DDIM~\cite{song2022}, PNDM~\cite{liu2022pseudo}, Stable Diffusion series (\textit{i.e.}, SDv1.5\_I2I, SDv1.5\_T2I, SDv2.1\_I2I, SDv2.1\_T2I)~\cite{rombach2021highresolution}, LDM ~\cite{rombach2021highresolution}, Inpaint~\cite{rombach2021highresolution}, and DiffSwap~\cite{zhao2023diffswap}. DiFF~\cite{Cheng2024} contains over 500,000 images synthesized by 13 advanced diffusion models across Face Edit, Face Swap, I2I, and T2I categories, including Midjourney~\cite{midjourney}, SDXL (\textit{i.e.}, SDXL, SDXL\_Refine)~\cite{podell2023sdxlimprovinglatentdiffusion}, FreeDoM (\textit{i.e.}, FreeDoM\_I, FreeDoM\_T)~\cite{yu2023freedom}, HPS~\cite{wu2023better}, LoRA~\cite{hu2022lora}, DreamBooth~\cite{ruiz2023dreambooth}, DiffFace~\cite{kim2022diffface}, DCFace~\cite{kim2023dcfacesyntheticfacegeneration}, Imagic~\cite{kawar2023imagictextbasedrealimage}, CycleDiff~\cite{cyclediffusion}, and CoDiff~\cite{huang2023collaborative}.

\paragraph{Implementation Details.}
We utilize the publicly available ADM checkpoint~\cite{dhariwal2021diffusionmodelsbeatgans} as diffusion noise estimation network and set the timestep $t=1$. The forensic classifier is based on a ResNet-50~\cite{he2015deepresiduallearningimage} architecture. We use the Adam optimizer with a learning rate of $1 \times 10^{-5}$ and a batch size of 64 for training. Each model is trained for 20 epochs. All experiments are conducted on a single NVIDIA GeForce RTX 2080 Ti GPU.

\begin{table*}[!t] 
  \centering 
  \centering \setlength{\tabcolsep}{22pt}\small
  \caption{Effect of different diffusion models on performance evaluated on DiFF. Models marked with $^{*}$ denote those retrained with the training dataset in our work. We report the average ACC and AP via cross-model evaluation on Diff. The best and second-best results are marked in bold and underline.} 
  \label{tab:noise_model_performance} 
  \begin{tabular}{|r|c|c|c||cc|}
\hline\thickhline
     \rowcolor{mygray}
    \textbf{ANL model} & \textbf{Training dataset} & \textbf{Image size} & \textbf{Conditional} & {\textbf{ACC}} & \textbf{AP} \\ 
    \hline
    iDDPM~\cite{nichol2021improveddenoisingdiffusionprobabilistic} & ImageNet~\cite{russakovsky2015imagenetlargescalevisual} & 64$\times$64 & \ding{51} & 54.28&68.89 \\
    iDDPM~\cite{nichol2021improveddenoisingdiffusionprobabilistic} & ImageNet~\cite{russakovsky2015imagenetlargescalevisual} & 64$\times$64 & \ding{55} & 57.20&74.22 \\ 
    
    iDDPM$^{*}$ ~\cite{nichol2021improveddenoisingdiffusionprobabilistic}& FFHQ~\cite{karras2019stylebasedgeneratorarchitecturegenerative} & 256$\times$256 & \ding{55}  & 62.07 &67.75 \\ 
    iDDPM~\cite{nichol2021improveddenoisingdiffusionprobabilistic} & ImageNet~\cite{russakovsky2015imagenetlargescalevisual} & 256$\times$256 & \ding{55} & 64.23 & \underline{88.91}\\ 
    ADM~\cite{dhariwal2021diffusionmodelsbeatgans} & ImageNet~\cite{russakovsky2015imagenetlargescalevisual} & 64$\times$64 & \ding{51}   & 59.51&76.29\\
    ADM~\cite{dhariwal2021diffusionmodelsbeatgans} & ImageNet~\cite{russakovsky2015imagenetlargescalevisual} & 64$\times$64 & \ding{55}  & 61.71 & 80.23\\
    
    ADM$^{*}$~\cite{dhariwal2021diffusionmodelsbeatgans} & FFHQ~\cite{karras2019stylebasedgeneratorarchitecturegenerative} & 256$\times$256 & \ding{55}   & \underline{66.30} &73.21\\
    ADM~\cite{dhariwal2021diffusionmodelsbeatgans} & ImageNet~\cite{russakovsky2015imagenetlargescalevisual} & 256$\times$256 & \ding{55} & \textbf{72.53}&\textbf{92.52} \\
    \hline
  \end{tabular}
\vspace{-3mm}
\end{table*}
\paragraph{Performance Metrics.}
Following prior deepfake detection methods~\cite{Wang2023, ojha2024universalfakeimagedetectors, frank2020leveragingfrequencyanalysisdeep, wang2020cnngeneratedimagessurprisinglyeasy}, we report both accuracy (ACC) and average precision (AP) in our experiments to evaluate the detectors. 
For cross-model evaluation, besides reporting cross-model performance, we also calculate average ACC and AP to demonstrate the generalization of our method.

\subsection{Cross-model Evaluation}
\label{sec:cross_model}
To evaluate the generalization capability of our method on unseen generative models, we conducted cross-model evaluations on the DiffFace and DiFF datasets. We performed experiments using the official implementations of ResNet~\cite{he2015deepresiduallearningimage} and DIRE~\cite{Wang2023}, and compared their results with our proposed ANL method.
Due to space constraints, full results are provided in Appendix . Figure~\ref{fig:cross_model_heatmap} shows the ACC results of DIRE and ANL on DiffFace.

Based on the cross-model performance visualization in Figure~\ref{fig:cross_model_heatmap}, we have the following observations:
(1) Methods specifically designed for diffusion models, such as DIRE, still exhibit limited generalization when facing unseen generative models.
(2) Our proposed ANL method achieves the best results across most test scenarios, demonstrating strong cross-model generalization capabilities.

Furthermore, we extended our cross-model evaluation strategy by comparing our ANL method with existing deepfake detection methods, reporting the average results in Table~\ref{tab:cross-model_performance}. Our method significantly outperformed all other methods, achieving substantial ACC/AP improvements of at least 12.33\%/6.87\% on DiffFace and 4.51\%/9.81\% on DiFF.
These results consistently demonstrate the strong generalization capability of our method operating in the noise domain, owing to the fundamental differences between real and diffusion-generated images.

\subsection{Cross-dataset Evaluation}
\label{sec:cross_dataset}
In addition to cross-model validation, we perform cross-dataset evaluation to further demonstrate the generalization of our method. 
In this scenario, models are trained on one complete dataset and evaluated on another. 
We report results of various advanced methods in Table \ref{tab:cross-dataset_performance}. The direction of the arrow ($\rightarrow$) indicates the testing dataset.
Results indicate that most existing methods struggle to achieve effective generalization, typically obtaining accuracies below 70\%, due to overfitting to dataset-specific characteristics.
In contrast, our ANL operates in the noise domain, effectively minimizing interference from dataset-specific semantic information and thereby achieving superior cross-dataset generalization.
These results further show that ANL effectively leverages noise-domain analysis to achieve robust generalization across diverse datasets.

\subsection{Standard Evaluation}
\label{sec:standard}
In this work, we also evaluate ANL under the standard setting, \textit{i.e.}, training and testing on the same dataset.
We further test on a diverse dataset, DiffusionForensics~\cite{Wang2023}. 
From the results in Table \ref{tab:performance}, it can be observed that ANL achieves best or second-best performance across all datasets, demonstrating its effectiveness. However, this evaluation does not sufficiently reflect the generalization capabilities of the methods, since all methods achieve competitive performance. This highlights the importance of cross-dataset or cross-model evaluations for measuring generalization.

\subsection{Alation Study}
\label{sec:ablation}
\textbf{Ablation of Noise Estimation Model.}
The diffusion model used for noise estimation is a critical component of our ANL method. 
We compare the average cross-model ACC on DiFF using different noise estimation models, which vary in model, training dataset, image size and condition.
We present the results in Table~\ref{tab:noise_model_performance}. Models marked with $^{*}$ denote those retrained in this work. The remaining models are obtained from publicly available repositories~\cite{guided-diffusion, improved-diffusion}. 
The results demonstrate that the unconditional ADM model trained on the ImageNet dataset at a resolution of 256$\times$256 yields the best performance.
This superior performance can be attributed to the stronger capability of powerful diffusion models to simulate noise patterns and capture subtle differences between real and fake images.
In our work, we adopt this ADM model as our noise estimation network for all experiments.

\begin{figure}
    \centering
    \includegraphics[width=0.48\textwidth]{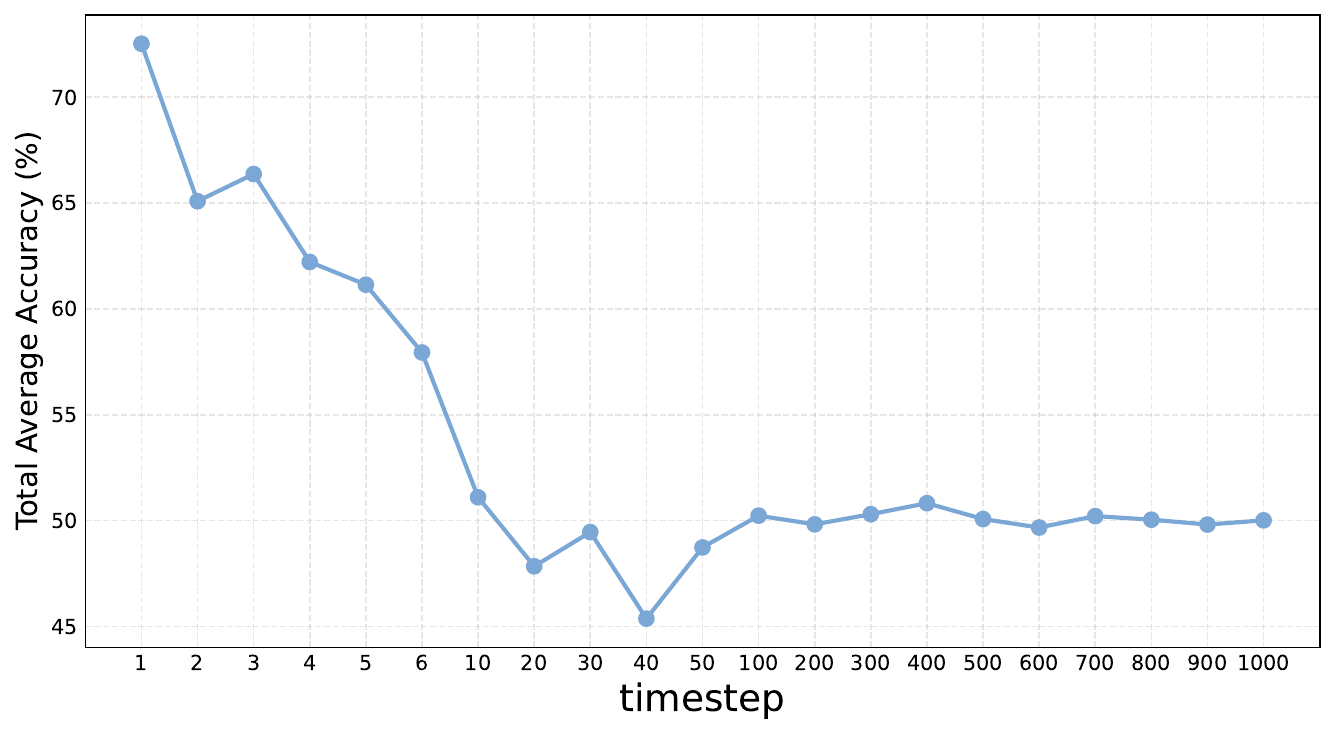}
    \caption{Effect of the timestep $t$ on cross-model generalization performance evaluated on DiFF.}
    \label{fig:timestep}
\end{figure}

\begin{figure}
    \centering 
    \includegraphics[width=0.48\textwidth]{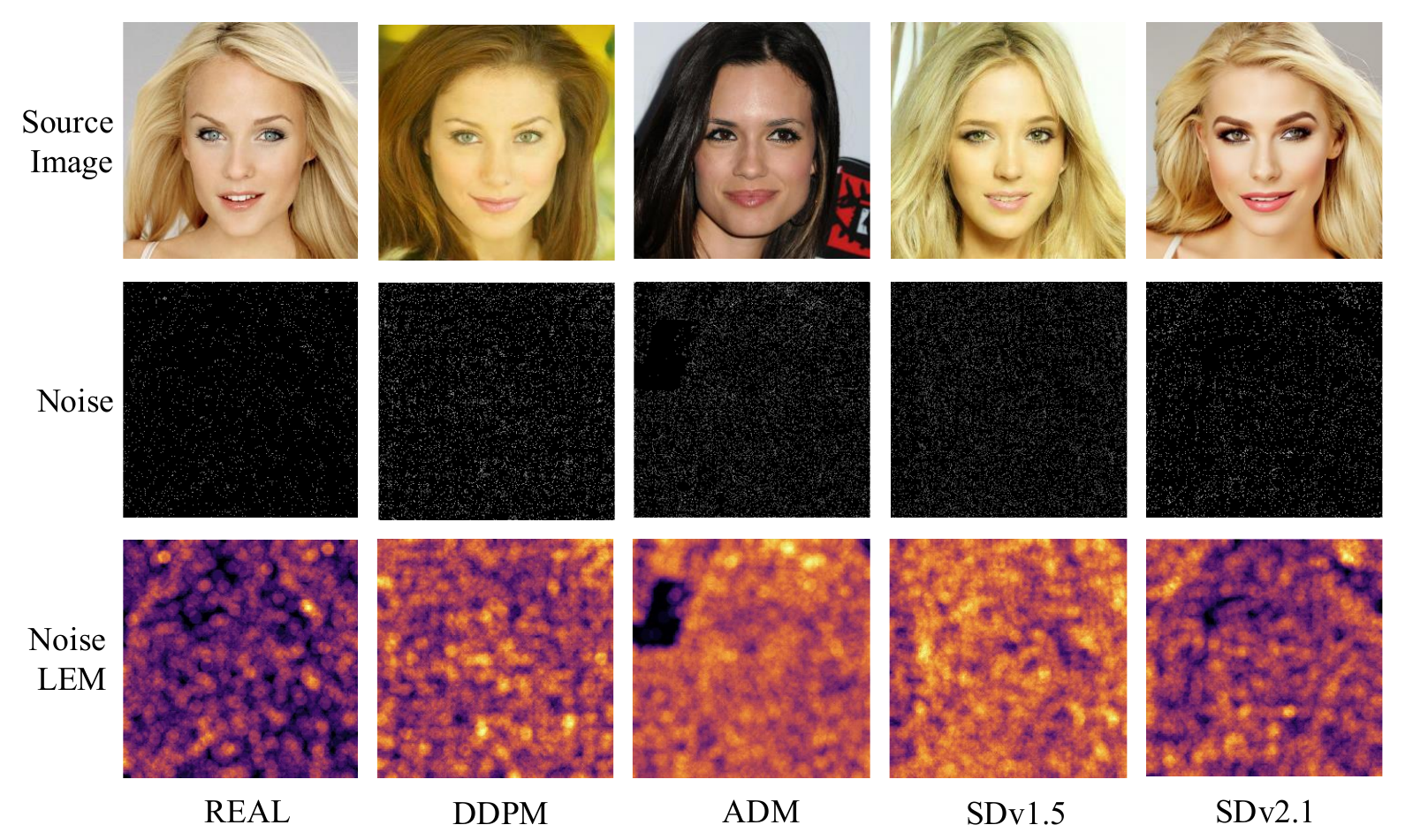} 
    \caption{Visualization of predicted noise and corresponding noise local entropy maps (LEM). We compare the predicted noise from real images and various diffusion-generated (fake) images. Compared to real image noise pattern, the generated image noise pattern is closer to white noise.} 
    \label{fig:LEM_visualization} 
\end{figure}

\begin{table*}
\centering \setlength{\tabcolsep}{16pt}\small
    \caption{Effect of the attention map. We report the average ACC and AP via cross-model and cross-dataset evaluation. The best and second-best results are marked in bold and underline.}\label{tab:alation-attention}
\begin{tabular}{|r||cc|cc|cc|cc|}
\hline\thickhline
\rowcolor{mygray}& \multicolumn{4}{c|}{Cross-Model} & \multicolumn{4}{c|}{Cross-Dataset}\\
\rowcolor{mygray}& \multicolumn{2}{c|}{DiffFace} & \multicolumn{2}{c|}{DiFF} & \multicolumn{2}{c|}{DiffFace $\rightarrow$ DiFF} & \multicolumn{2}{c|}{DiFF $\rightarrow$ DiffFace} \\
\rowcolor{mygray}\multirow{-3}[-1]{*}{Method}  & ACC             & AP    & ACC             & AP & ACC             & AP           & ACC             & \multicolumn{1}{c|}{AP}              \\
\hline\hline
\textbf{ANL (w/o AM)}           &77.55&89.02&70.13&80.14&69.22 & 79.81 &79.43& 88.34\\
\textbf{ANL (full)}           & \textbf{80.38}         & \textbf{95.61}        & \textbf{72.53}         & \textbf{92.52} & \textbf{82.61}         & \textbf{93.27}       & \textbf{92.96}         & \textbf{96.22}          \\
\hline
\end{tabular}
\end{table*}
\vspace{-4mm}
\textbf{Effect of the Timestep $t$.}
The denoising process of diffusion models naturally exhibits a coarse-to-fine characteristic: at larger timesteps (early stages of the denoising process), the model primarily controls coarse, global image structures, whereas at smaller timesteps (later stages), it progressively refines fine-grained details and subtle textures
~\cite{patashnik2023localizing, voynov2023P+}. In this section, we systematically evaluate the generalization performance by cross-model evaluations (timestep $t=1$ to $t=1000$) on DiFF. The results are shown in Figure~\ref{fig:timestep}, with detailed numerical results provided in Appendix.
Experimental results show that ANL achieves optimal generalization performance at the smallest timestep ($t=1$). This observation aligns well with the design rationale of ANL. As the timestep decreases, noise  more clearly capture subtle yet discriminative artifacts, directly contributing to enhanced cross-model generalization.

\textbf{Effect of the Attention Map.}
We further ablate the proposed attention-guided design by removing attention map and directly feeding the predicted noise into the classifier. The results in Table~\ref{tab:alation-attention} show that attention map consistently improves performance under both cross-model and cross-dataset settings. A particularly larger gain in cross-dataset scenario, achieving ACC/AP improvements of 13.39\%/13.46\% on DiffFace to Diff and
13.53\%/7.88\% on DiFF to DiffFace. This suggests that the benefit of attention mechanism goes beyond simple feature enhancement and is closely related to generalization.
Without the attention map, the classifier tends to rely more on local patterns in the predicted noise, which are more sensitive to dataset-specific biases (\textit{e.g.}, image quality, cropping style, compression, and identity distribution). In contrast, proposed noise-derived attention map explicitly highlights the global spatial distribution of noise intensity, guiding the backbone to focus on more stable and general forensic cues.

\vspace{-2mm}
\subsection{Visualizations of Predicted Noise and Noise LEM}
\label{sec:visualization}
To intuitively demonstrate the feasibility and effectiveness of our ANL operating in the diffusion noise domain, we visualize the predicted noise and corresponding noise local entropy maps (LEM) in Figure ~\ref{fig:LEM_visualization}.
The first row shows source images: one real image and several representative diffusion-generated images.
The second row presents noise estimations from these images by our ANL.
Notably, noise predicted from real images distinctly captures richer fine-grained details, whereas that from diffusion-generated images closely resembles featureless white noise.

To make the contrast even more pronounced, we further visualize LEM of the predicted noise (third row), where per-patch entropy calculations clearly differentiate real images from diffusion-generated ones.
The distinctiveness captured by these noise patterns and their entropy distributions demonstrates ANL's capability to effectively identify subtle yet informative differences between real and fake images.
Additionally, we observe that noise representations from more advanced diffusion models (\textit{e.g.}, SDv2.1) closely resemble those of real images. 
This indicates that current SOTA diffusion models can effectively replicate noise details inherent in real images, thus making generated images increasingly realistic. Nevertheless, significant discrepancies in noise patterns remain evident, underscoring the feasibility of distinguishing diffusion-generated images from real ones.

\vspace{-2mm}
\section{Conclusion}
This paper introduces Attention-guided Noise Learning (ANL), a principled approach to identifying diffusion-generated deepfakes by shifting the forensic focus to the noise domain. Unlike conventional pixel-based methods, ANL harnesses high-fidelity noise priors from pre-trained diffusion models to expose the subtle structural and statistical irregularities inherent in synthetic generation. Our mechanism effectively bridges the "generalization gap" by isolating model-agnostic artifacts that remain invariant across different generative architectures.
Beyond the methodology, we contribute a rigorous cross-model evaluation strategy that simulates real-world scenarios. Comprehensive experimental resultsvalidate that ANL not only achieves superior detection performance but also exhibits remarkable stability against unseen generators. We believe that both ANL framework and our evaluation methodology offer a critical foundation for future research, moving the community toward more transparent and robust deepfake detection systems.

\nocite{*}

\newpage
\bibliographystyle{ACM-Reference-Format}
\bibliography{MM26}
\newpage
\appendix
\section{Research Methods}

\subsection{Part One}
\section*{Overview of Appendix}
The appendix is organized as follows:
\begin{itemize}
     \item \textbf{Appendix~{\ref{appendix:cross_model}}: Cross-model Evaluation Results.}
     \item \textbf{Appendix~{\ref{timestep}}: Effect of the Timestep $t$ on DiFF.}
\end{itemize}

\section{Cross-model Evaluation Results}
\begin{figure*}[htbp]
  \centering
  \begin{minipage}[t]{0.485\textwidth}
    \centering
    \includegraphics[width=\textwidth]{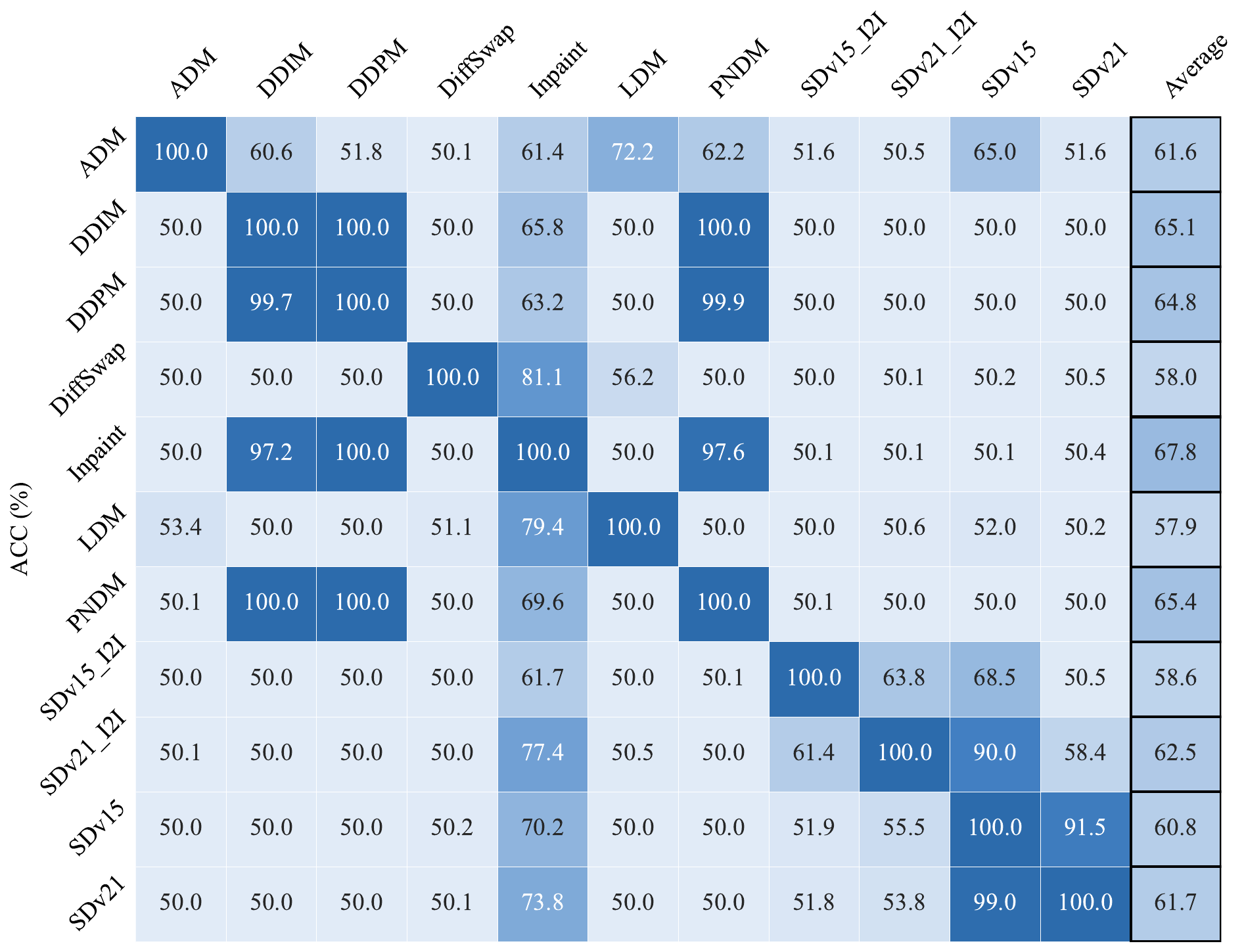}
    \smallskip
  \end{minipage}
  \hfill
  \begin{minipage}[t]{0.485\textwidth}
    \centering
    \includegraphics[width=\textwidth]{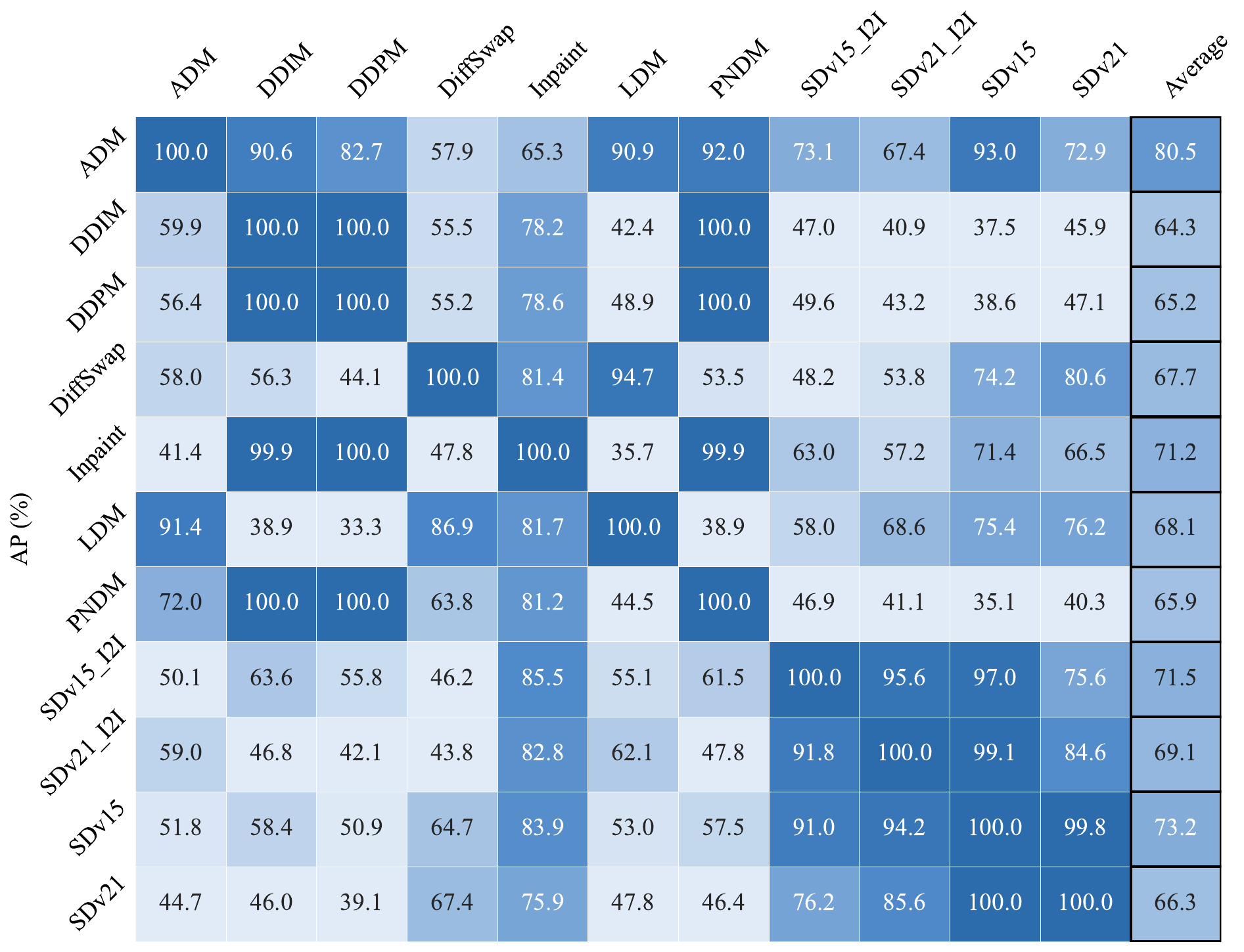}
    \smallskip
  \end{minipage}\\[-10pt]
  \textbf{(a)} ResNet on DiffFace
  
    \begin{minipage}[t]{0.485\textwidth}
    \centering
    \includegraphics[width=\textwidth]{./figures/dire_heatmap_ACC.pdf}
    \smallskip
  \end{minipage}
  \hfill
  \begin{minipage}[t]{0.485\textwidth}
    \centering
    \includegraphics[width=\textwidth]{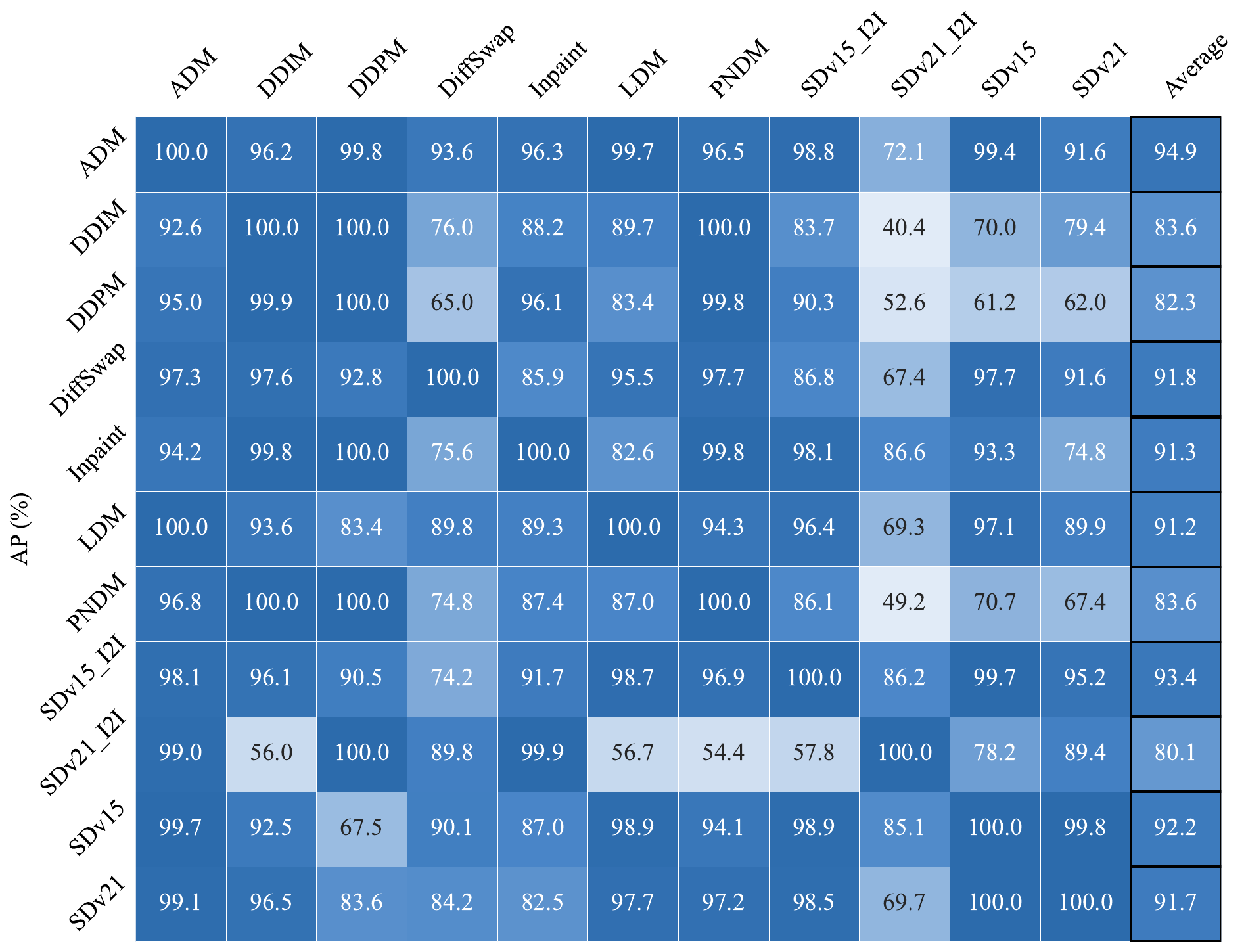}
    \smallskip
  \end{minipage}\\[-10pt]
  \textbf{(b)} DIRE on DiffFace

    \begin{minipage}[t]{0.485\textwidth}
    \centering
    \includegraphics[width=\textwidth]{./figures/NP_heatmap_ACC.pdf}
    \smallskip
  \end{minipage}
  \hfill
  \begin{minipage}[t]{0.485\textwidth}
    \centering
    \includegraphics[width=\textwidth]{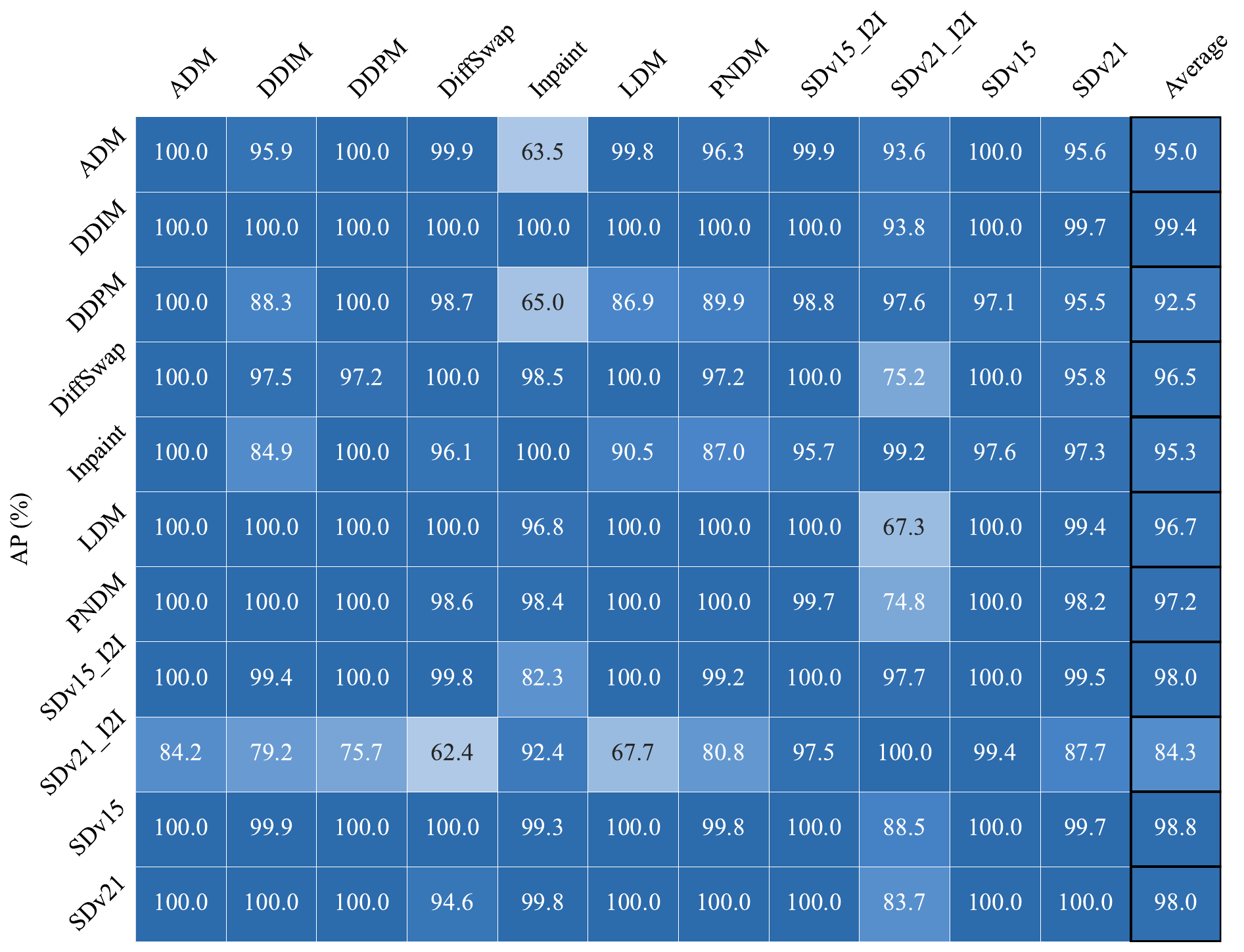}
    \smallskip
  \end{minipage}\\[-10pt]
    \textbf{(c)} ANL on DiffFace

  \caption{Cross-model evaluation results (ACC and AP) on DiffFace. Vertical comparison of results.
}
  \label{fig:cross_model_heatmap_diffface}
\end{figure*}

\begin{figure*}[htbp]
\vspace{-3mm}
  \centering
  \begin{minipage}[t]{0.485\textwidth}
    \centering
    \includegraphics[width=\textwidth]{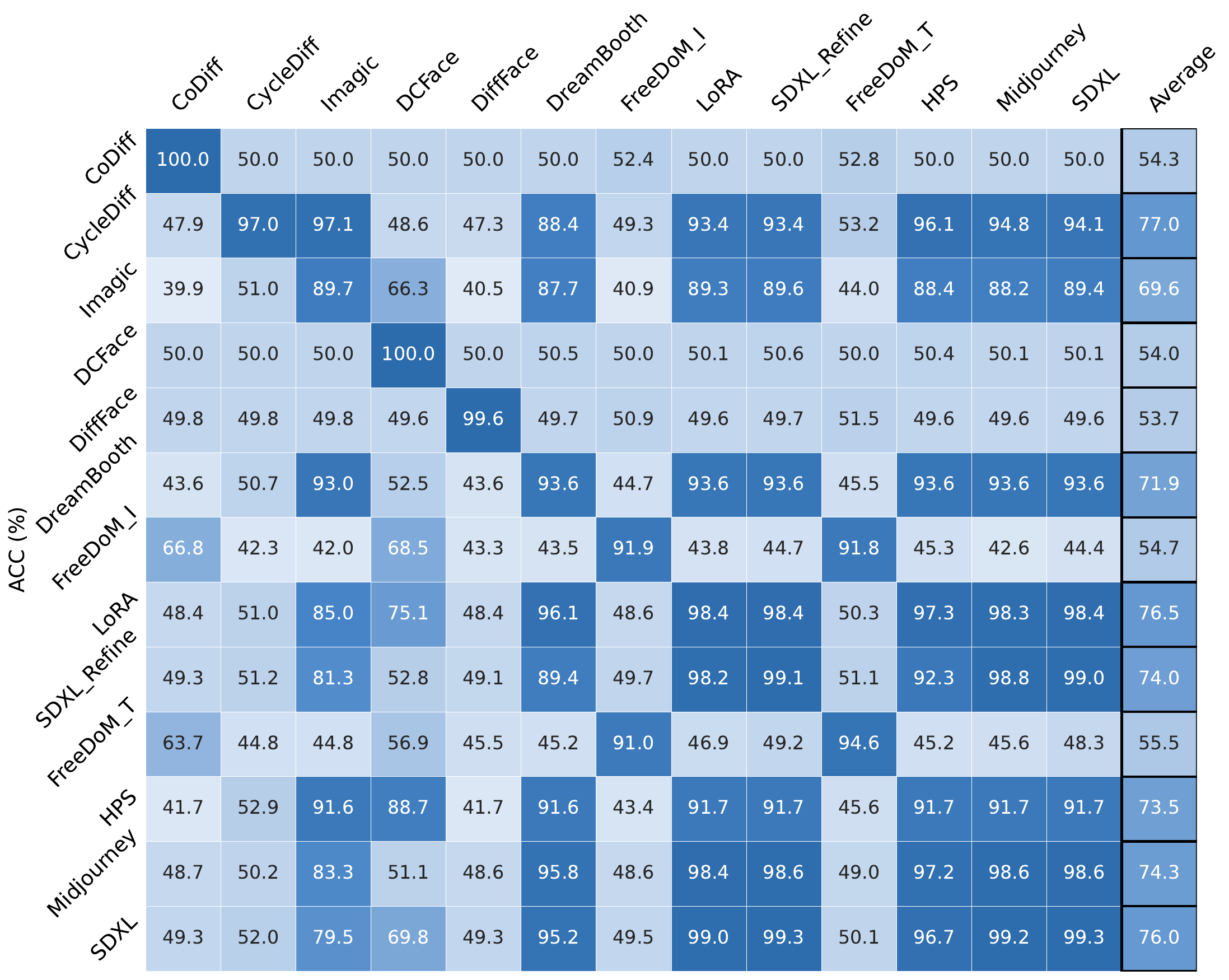}
    \smallskip
  \end{minipage}
  \hfill
  \begin{minipage}[t]{0.485\textwidth}
    \centering
    \includegraphics[width=\textwidth]{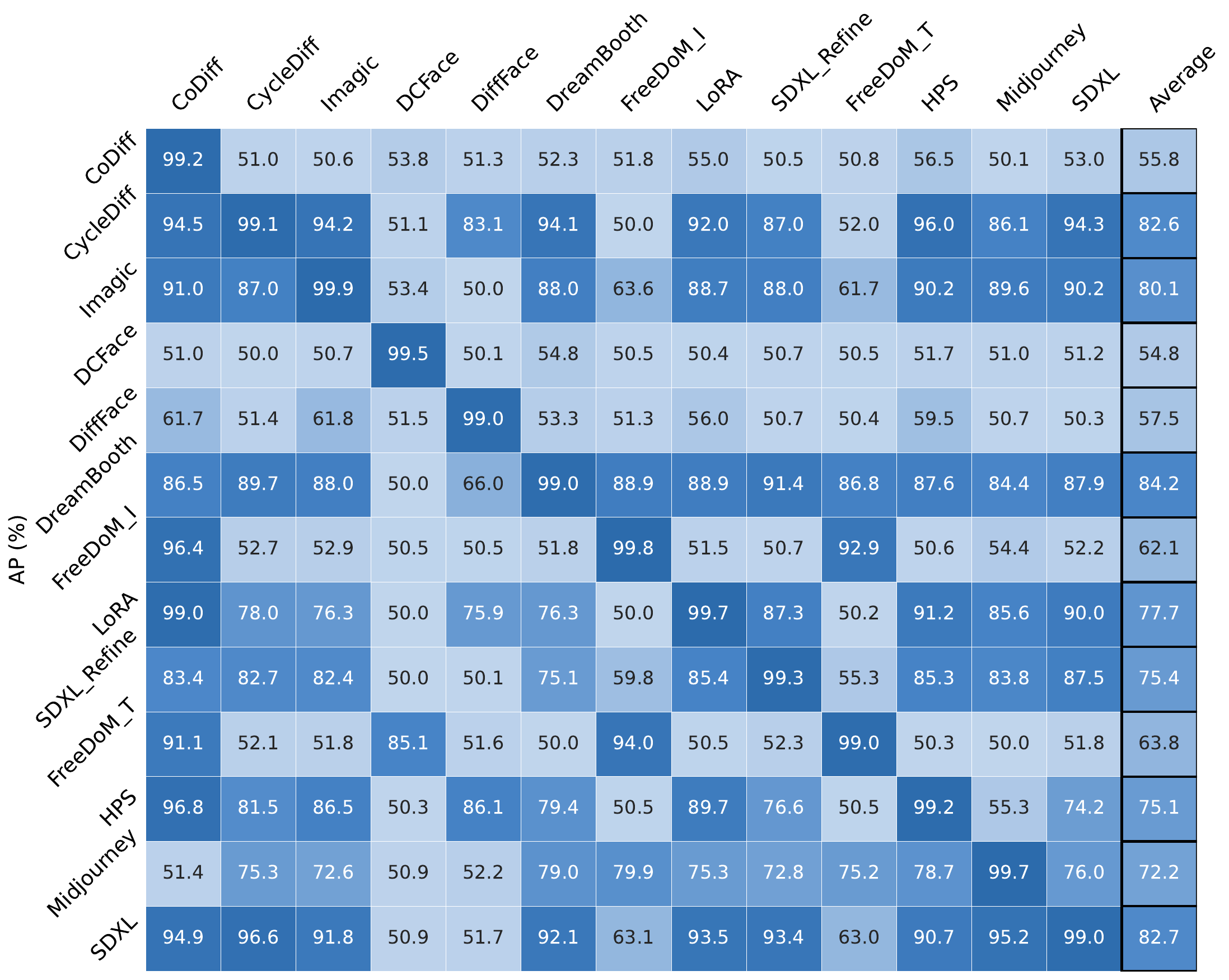}
    \smallskip
  \end{minipage}\\[-10pt]
  \textbf{(a)} ResNet on DiFF
  
    \begin{minipage}[t]{0.485\textwidth}
    \centering
    \includegraphics[width=\textwidth]{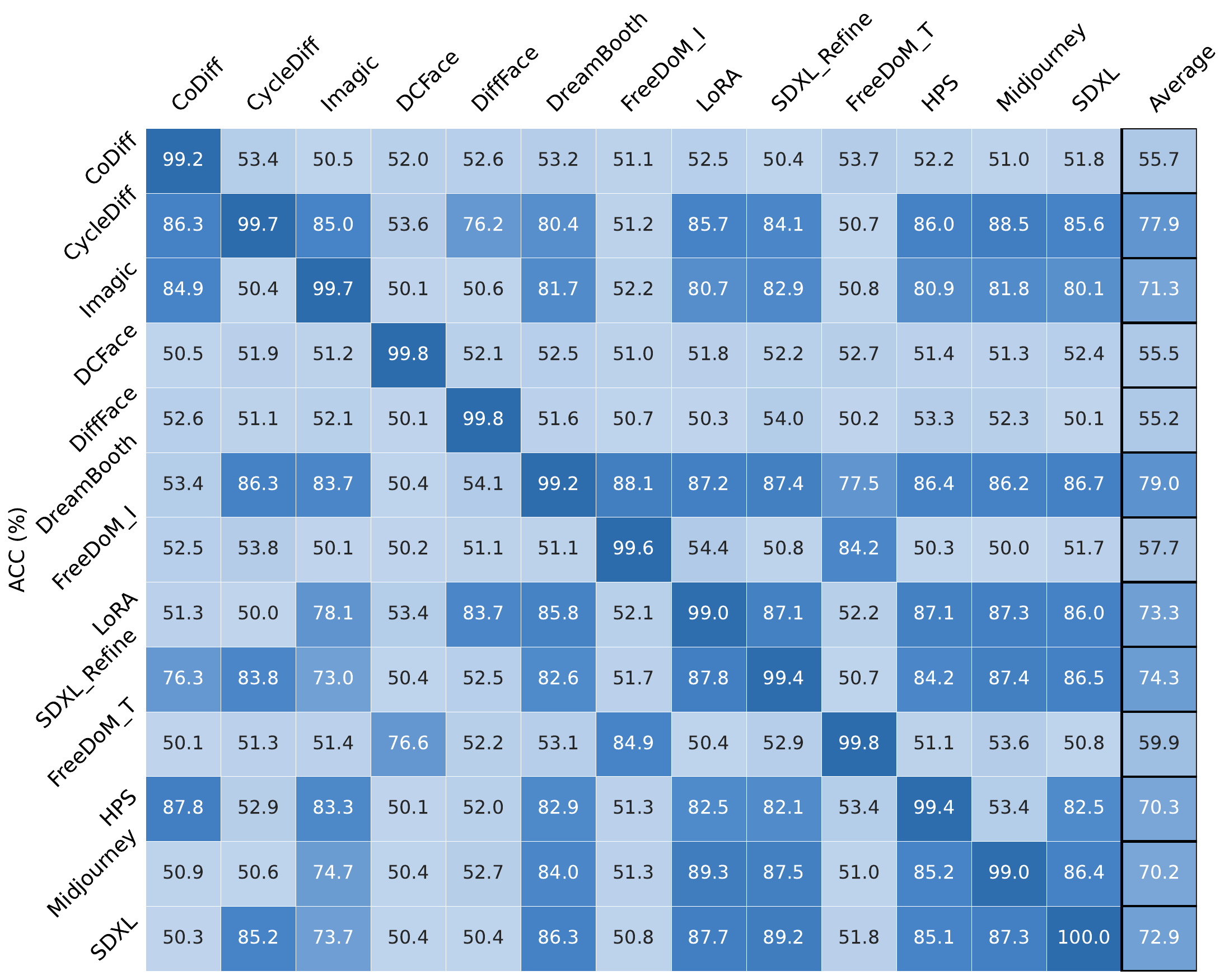}
    \smallskip
  \end{minipage}
  \hfill
  \begin{minipage}[t]{0.485\textwidth}
    \centering
    \includegraphics[width=\textwidth]{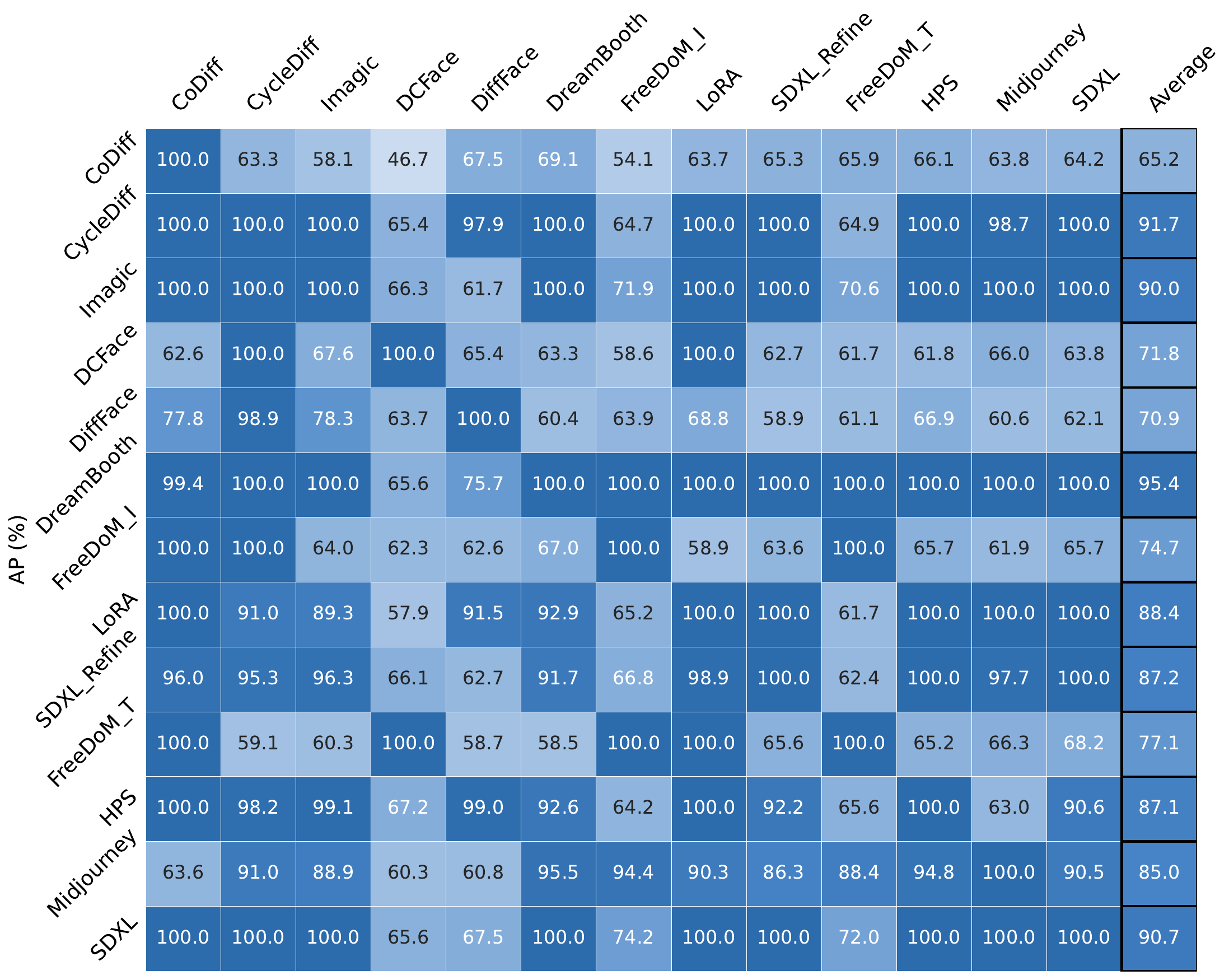}
    \smallskip
  \end{minipage}\\[-10pt]
  \textbf{(b)} DIRE on DiFF

    \begin{minipage}[t]{0.485\textwidth}
    \centering
    \includegraphics[width=\textwidth]{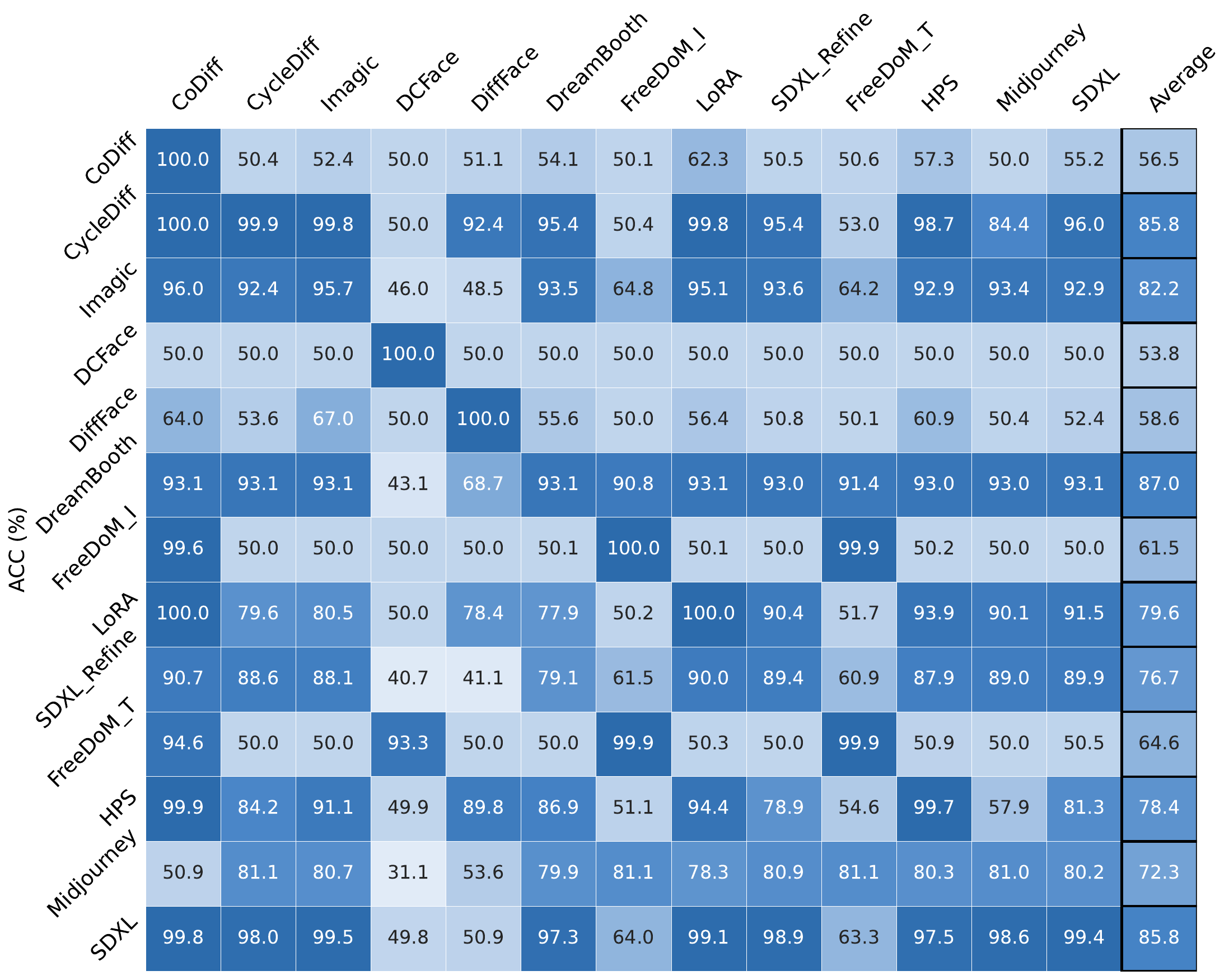}
    \smallskip
  \end{minipage}
  \hfill
  \begin{minipage}[t]{0.485\textwidth}
    \centering
    \includegraphics[width=\textwidth]{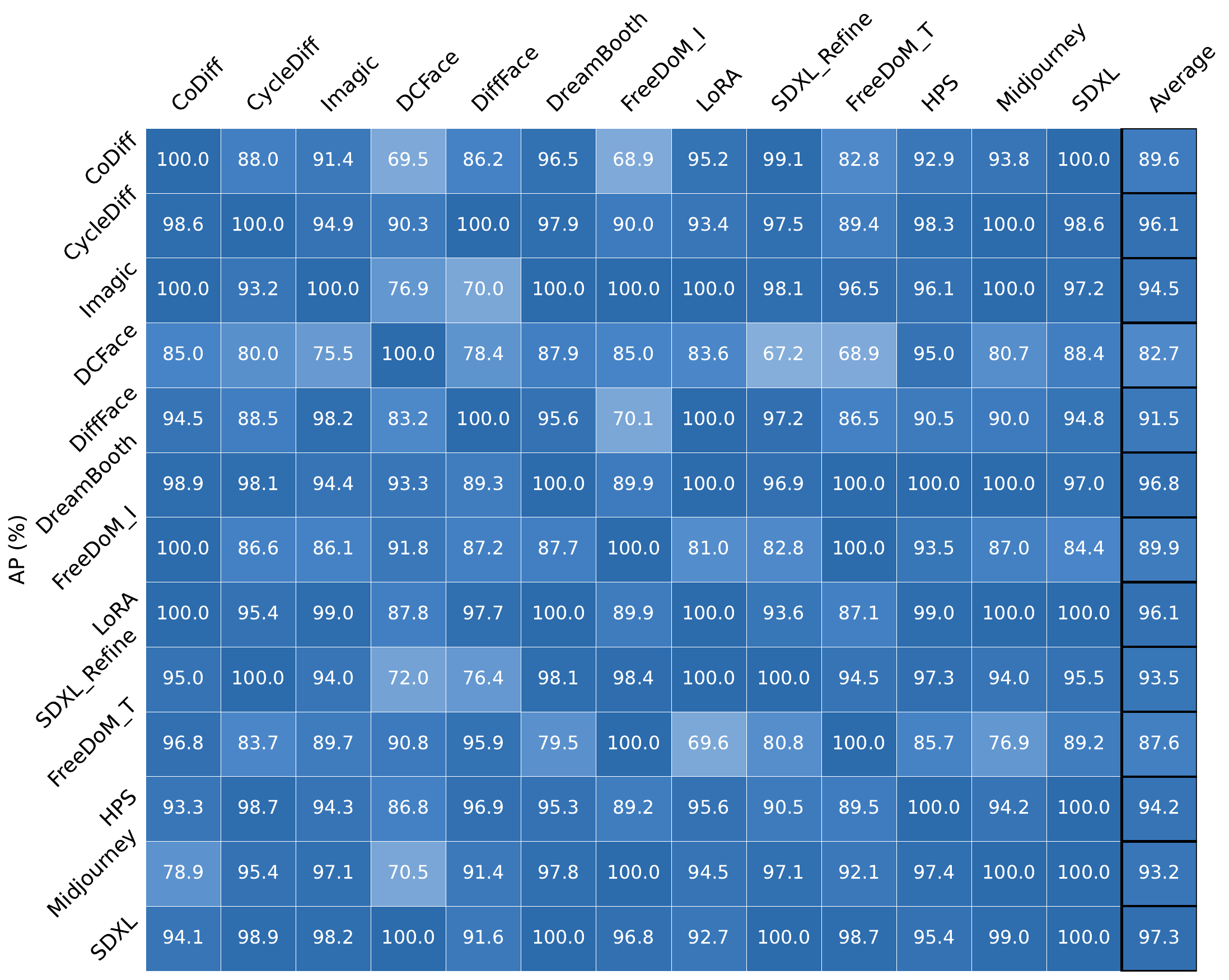}
    \smallskip
  \end{minipage}\\[-10pt]
    \textbf{(c)} ANL on DiFF
  \vspace{-3mm}

  \caption{Cross-model evaluation results (ACC and AP) on DiFF. Vertical comparison of results.}
  \label{fig:cross_model_heatmap_diff}
\end{figure*}

To thoroughly demonstrate the strong generalization capability of our ANL, we conduct full cross-model generalization experiments on DiFF and DiffFace, with the experimental results presented in Figure \ref{fig:cross_model_heatmap_diffface} and Figure \ref{fig:cross_model_heatmap_diff}. The labels on the left indicate the generative models used for training (\textit{i.e.}, data generated by these models), while the labels at the top represent the generative models used for testing. Darker colors correspond to better performance.

We select CNN-based models (\textit{e.g.}, ResNet) and DIRE for comparison with our proposed ANL. Specifically, ResNet, as a direct classification approach, serves as our baseline, while DIRE is an advanced deepfake detection method. The experimental results indicate that CNN-based methods exhibit poor generalization. Although DIRE demonstrates effectiveness, it still underperforms in terms of generalization compared to our ANL. These comparisons clearly illustrate that our method, which operates in the diffusion noise domain, achieves strong generalization capabilities across unseen generative models.

\label{appendix:cross_model}

\section{Effect of the Timestep $t$ on DiFF}
\label{timestep}
Table~\ref{tab:timestep1to4}, Table~\ref{tab:timestep5to20}, Table~\ref{tab:timestep3to10}, Table~\ref{tab:timestep20to80} and Table~\ref{tab:timestep100} report the ACC of cross-model evaluation with ADM (256$\times$256 diffusion not class conditional) from timestep 1 to 1000. 
We conduct experiments at several key steps. 
Experimental results indicate that smaller timesteps yield better performance, and we provide an analysis of the reasons in the main paper.

\begin{table*}[htbp] 
\centering
\centering \setlength{\tabcolsep}{3pt}\small
\caption{Cross-model ACC on DiFF (timestep $t$ from 1 to 4).}
\label{tab:timestep1to4}
\begin{adjustbox}{max width=\textwidth}
\begin{tabular}{|l| *{14}{|r}|}
\hline\thickhline
     \rowcolor{mygray}
$t=1$ & CoDiff & cycle\_diff & Imagic & DCFace & DiffFace & DreamBooth & FreeDoM\_I & LoRA & SDXL\_Refine & FreeDoM\_T & HPS & Midjourney & SDXL & real \\
\hline\hline
CoDiff & 100.00 & 0.05 & 0.00 & 0.09 & 0.00 & 0.00 & 4.87 & 0.00 & 0.02 & 5.64 & 0.01 & 0.00 & 0.02 & 100.00 \\
cycle\_diff & 1.32 & 99.65 & 99.75 & 2.70 & 0.13 & 82.44 & 4.25 & 92.40 & 92.31 & 12.03 & 97.68 & 95.13 & 93.81 & 94.43 \\
Imagic & 0.25 & 22.51 & 99.88 & 53.08 & 1.59 & 95.86 & 2.39 & 99.18 & 99.80 & 8.62 & 97.28 & 96.95 & 99.42 & 79.45 \\
DCFace & 0.00 & 0.00 & 0.00 & 100.00 & 0.00 & 1.00 & 0.00 & 0.13 & 1.12 & 0.00 & 0.70 & 0.12 & 0.20 & 100.00 \\
DiffFace & 0.42 & 0.38 & 0.42 & 0.00 & 100.00 & 0.18 & 2.57 & 0.00 & 0.16 & 3.85 & 0.04 & 0.02 & 0.04 & 99.17 \\
DreamBooth & 0.07 & 14.21 & 98.85 & 17.91 & 0.00 & 99.95 & 2.22 & 100.00 & 100.00 & 3.77 & 99.96 & 99.95 & 100.00 & 87.17 \\
FreeDoM\_I & 49.78 & 0.84 & 0.25 & 53.25 & 2.79 & 3.27 & 100.00 & 3.71 & 5.64 & 99.72 & 6.77 & 1.35 & 5.05 & 83.79 \\
LoRA & 0.05 & 5.27 & 73.22 & 53.45 & 0.00 & 95.45 & 0.40 & 100.00 & 100.00 & 3.73 & 97.83 & 99.86 & 99.96 & 96.81 \\
SDXL\_Refine & 0.37 & 4.10 & 64.40 & 7.32 & 0.00 & 80.53 & 1.11 & 98.12 & 100.00 & 3.85 & 86.44 & 99.40 & 99.80 & 98.26 \\
FreeDoM\_T & 38.08 & 0.35 & 0.28 & 24.45 & 1.71 & 1.05 & 92.78 & 4.46 & 9.02 & 99.96 & 1.12 & 1.82 & 7.28 & 89.29 \\
HPS & 0.02 & 22.40 & 99.86 & 94.10 & 0.06 & 99.80 & 3.32 & 100.00 & 100.00 & 7.78 & 100.00 & 100.00 & 100.00 & 83.38 \\
Midjourney & 0.05 & 3.06 & 69.27 & 4.84 & 0.00 & 94.30 & 0.00 & 99.56 & 99.96 & 0.68 & 97.05 & 99.98 & 99.89 & 97.27 \\
SDXL & 0.12 & 5.47 & 60.47 & 41.13 & 0.00 & 91.90 & 0.44 & 99.56 & 100.00 & 1.59 & 94.83 & 99.96 & 100.00 & 98.52 \\
\hline
\hline\thickhline
     \rowcolor{mygray}
$t=2$ & CoDiff & cycle\_diff & Imagic & DCFace & DiffFace & DreamBooth & FreeDoM\_I & LoRA & SDXL\_Refine & FreeDoM\_T & HPS & Midjourney & SDXL & real \\
\hline\hline
CoDiff & 99.98 & 0.01 & 0.21 & 100.00 & 0.00 & 0.59 & 0.18 & 4.15 & 0.05 & 0.56 & 1.85 & 0.00 & 3.43 & 100.00 \\
cycle\_diff & 77.99 & 4.99 & 0.60 & 0.00 & 99.94 & 1.41 & 0.00 & 22.30 & 0.23 & 0.04 & 10.38 & 0.87 & 6.06 & 99.99 \\
Imagic & 95.97 & 50.29 & 97.18 & 99.44 & 84.84 & 83.87 & 0.04 & 98.74 & 66.60 & 0.00 & 91.72 & 79.94 & 79.08 & 99.46 \\
DCFace & 0.00 & 0.00 & 0.00 & 100.00 & 0.00 & 0.00 & 0.00 & 0.00 & 0.00 & 0.00 & 0.00 & 0.00 & 0.00 & 100.00 \\
DiffFace & 0.00 & 0.54 & 0.12 & 0.00 & 100.00 & 0.05 & 0.00 & 1.82 & 0.00 & 0.00 & 0.48 & 0.11 & 0.38 & 100.00 \\
DreamBooth & 88.21 & 31.66 & 98.75 & 99.98 & 90.48 & 95.37 & 0.13 & 99.75 & 87.32 & 0.91 & 98.41 & 96.55 & 94.79 & 85.76 \\
FreeDoM\_I & 99.85 & 0.71 & 0.02 & 0.00 & 0.13 & 0.74 & 100.00 & 0.88 & 2.51 & 100.00 & 3.03 & 3.68 & 1.71 & 81.18 \\
LoRA & 100.00 & 66.81 & 96.19 & 100.00 & 99.49 & 90.72 & 0.97 & 99.94 & 91.67 & 2.82 & 99.25 & 97.08 & 94.75 & 96.82 \\
SDXL\_Refine & 96.02 & 59.48 & 99.93 & 99.09 & 81.15 & 99.18 & 5.80 & 99.75 & 94.95 & 3.46 & 99.10 & 93.20 & 97.00 & 43.54 \\
FreeDoM\_T & 99.30 & 48.41 & 10.90 & 0.00 & 22.14 & 25.15 & 100.00 & 22.36 & 67.33 & 100.00 & 26.79 & 63.06 & 50.04 & 0.83 \\
HPS & 98.16 & 27.58 & 90.95 & 100.00 & 95.05 & 86.94 & 0.62 & 99.31 & 85.89 & 2.10 & 95.29 & 98.41 & 90.81 & 99.58 \\
Midjourney & 84.50 & 74.83 & 98.91 & 82.33 & 99.24 & 95.22 & 0.97 & 99.94 & 94.95 & 1.79 & 98.82 & 99.63 & 97.51 & 78.91 \\
SDXL & 88.61 & 16.67 & 97.62 & 100.00 & 51.71 & 98.65 & 26.01 & 99.94 & 99.36 & 30.50 & 99.09 & 98.87 & 99.58 & 46.80 \\
\hline
\hline\thickhline
     \rowcolor{mygray}
$t=3$ & CoDiff & cycle\_diff & Imagic & DCFace & DiffFace & DreamBooth & FreeDoM\_I & LoRA & SDXL\_Refine & FreeDoM\_T & HPS & Midjourney & SDXL & real \\
\hline
CoDiff & 99.70 & 0.01 & 0.00 & 100.00 & 0.00 & 0.00 & 0.18 & 0.44 & 0.00 & 0.56 & 0.23 & 0.00 & 0.82 & 100.00 \\
cycle\_diff & 77.04 & 96.64 & 87.25 & 0.00 & 100.00 & 72.55 & 1.51 & 93.66 & 57.00 & 0.68 & 79.60 & 92.81 & 66.61 & 92.37 \\
Imagic & 41.14 & 16.23 & 92.73 & 0.00 & 82.36 & 60.72 & 0.00 & 87.81 & 32.12 & 0.00 & 68.75 & 52.10 & 42.72 & 98.65 \\
DCFace & 0.00 & 0.00 & 0.00 & 100.00 & 0.00 & 0.00 & 0.00 & 0.00 & 0.00 & 0.00 & 0.00 & 0.00 & 0.00 & 100.00 \\
DiffFace & 0.07 & 0.87 & 0.58 & 0.00 & 100.00 & 0.46 & 0.00 & 4.59 & 0.00 & 0.00 & 1.60 & 2.50 & 0.54 & 99.99 \\
DreamBooth & 66.87 & 22.86 & 89.66 & 0.00 & 98.79 & 85.71 & 0.09 & 96.92 & 52.35 & 0.40 & 92.41 & 73.84 & 65.50 & 95.72 \\
FreeDoM\_I & 99.38 & 0.87 & 0.51 & 0.00 & 4.82 & 0.61 & 98.32 & 0.88 & 2.12 & 97.42 & 1.61 & 3.13 & 1.29 & 73.31 \\
LoRA & 99.08 & 33.30 & 93.07 & 92.16 & 100.00 & 90.49 & 0.75 & 99.94 & 91.97 & 1.87 & 97.28 & 98.65 & 94.95 & 91.76 \\
SDXL\_Refine & 54.88 & 8.61 & 22.84 & 0.02 & 0.44 & 22.18 & 0.00 & 73.99 & 18.66 & 0.16 & 46.26 & 36.46 & 25.20 & 99.96 \\
FreeDoM\_T & 5.85 & 0.94 & 0.25 & 0.00 & 0.00 & 0.00 & 42.18 & 0.00 & 0.00 & 40.39 & 0.01 & 0.00 & 0.00 & 99.36 \\
HPS & 99.90 & 13.87 & 61.60 & 100.00 & 97.40 & 55.18 & 1.37 & 80.65 & 31.50 & 2.62 & 85.39 & 17.58 & 43.92 & 94.54 \\
Midjourney & 1.22 & 38.62 & 90.14 & 99.01 & 100.00 & 96.91 & 15.24 & 99.62 & 98.95 & 15.17 & 97.76 & 99.93 & 99.20 & 64.24 \\
SDXL & 18.76 & 9.65 & 79.50 & 100.00 & 92.89 & 87.86 & 0.40 & 99.18 & 86.93 & 1.63 & 92.00 & 98.11 & 91.32 & 95.33 \\
\hline
\hline\thickhline
     \rowcolor{mygray}
$t=4$ & CoDiff & cycle\_diff & Imagic & DCFace & DiffFace & DreamBooth & FreeDoM\_I & LoRA & SDXL\_Refine & FreeDoM\_T & HPS & Midjourney & SDXL & real \\
\hline\hline
CoDiff & 99.93 & 0.07 & 0.00 & 0.00 & 0.25 & 0.05 & 0.35 & 0.13 & 0.00 & 2.70 & 0.30 & 0.00 & 0.38 & 99.98 \\
cycle\_diff & 2.61 & 80.16 & 52.07 & 0.00 & 100.00 & 32.56 & 0.00 & 67.96 & 5.87 & 0.04 & 44.74 & 54.13 & 15.63 & 99.35 \\
Imagic & 93.46 & 93.47 & 99.95 & 0.00 & 97.14 & 96.88 & 3.41 & 98.12 & 66.95 & 0.32 & 95.43 & 85.94 & 81.43 & 29.01 \\
DCFace & 0.00 & 0.00 & 0.00 & 100.00 & 0.00 & 0.00 & 0.00 & 0.00 & 0.00 & 0.00 & 0.00 & 0.00 & 0.00 & 100.00 \\
DiffFace & 0.00 & 0.94 & 1.39 & 0.00 & 99.94 & 0.89 & 0.00 & 3.33 & 0.04 & 0.00 & 1.41 & 1.42 & 0.34 & 100.00 \\
DreamBooth & 17.09 & 46.06 & 97.53 & 0.00 & 22.65 & 88.47 & 0.66 & 86.87 & 42.80 & 0.48 & 81.79 & 49.30 & 55.01 & 70.51 \\
FreeDoM\_I & 0.00 & 0.00 & 0.00 & 0.02 & 0.00 & 0.00 & 0.09 & 0.00 & 0.00 & 0.00 & 0.00 & 0.00 & 0.02 & 100.00 \\
LoRA & 37.34 & 11.66 & 64.44 & 50.85 & 99.94 & 80.68 & 0.35 & 98.74 & 89.03 & 1.27 & 90.80 & 94.97 & 90.20 & 94.06 \\
SDXL\_Refine & 0.70 & 0.61 & 11.61 & 100.00 & 85.41 & 26.30 & 1.82 & 77.83 & 69.88 & 4.81 & 52.39 & 71.03 & 66.36 & 99.29 \\
FreeDoM\_T & 4.05 & 0.00 & 0.00 & 0.50 & 0.00 & 0.00 & 43.64 & 0.00 & 0.00 & 56.12 & 0.00 & 0.00 & 0.00 & 99.99 \\
HPS & 48.33 & 11.01 & 64.77 & 70.81 & 98.79 & 69.18 & 0.04 & 95.04 & 46.10 & 0.00 & 84.00 & 79.02 & 58.77 & 99.66 \\
Midjourney & 2.31 & 16.06 & 74.60 & 35.10 & 98.79 & 91.69 & 11.48 & 99.25 & 97.70 & 13.58 & 94.27 & 99.43 & 97.37 & 75.16 \\
SDXL & 62.41 & 20.33 & 80.86 & 11.90 & 99.30 & 95.71 & 43.82 & 99.50 & 99.18 & 47.82 & 98.25 & 99.34 & 98.78 & 68.39 \\
\hline
\end{tabular}
\end{adjustbox}
\end{table*}

\begin{table*}[htbp] 
\centering
\centering \setlength{\tabcolsep}{3pt}\small
\caption{Cross-model ACC on DiFF (timestep $t$ from 5 to 20).}
\label{tab:timestep5to20}
\begin{adjustbox}{max width=\textwidth}
\begin{tabular}{|l| *{14}{|r}|}
\hline\thickhline
     \rowcolor{mygray}
$t=5$ & CoDiff & cycle\_diff & Imagic & DCFace & DiffFace & DreamBooth & FreeDoM\_I & LoRA & SDXL\_Refine & FreeDoM\_T & HPS & Midjourney & SDXL & real \\
\hline\hline
CoDiff & 99.83 & 0.14 & 0.00 & 0.00 & 0.44 & 0.00 & 1.24 & 0.44 & 0.05 & 3.89 & 0.15 & 0.00 & 0.24 & 99.87 \\
cycle\_diff & 76.92 & 99.99 & 99.61 & 98.19 & 100.00 & 95.40 & 43.42 & 99.50 & 84.80 & 32.57 & 97.25 & 99.13 & 90.70 & 7.17 \\
Imagic & 0.87 & 31.92 & 86.79 & 0.00 & 20.88 & 58.06 & 0.00 & 61.87 & 9.06 & 0.00 & 55.00 & 33.72 & 17.34 & 97.55 \\
DCFace & 0.00 & 0.00 & 0.00 & 100.00 & 0.00 & 0.00 & 0.00 & 0.00 & 0.00 & 0.00 & 0.00 & 0.00 & 0.00 & 100.00 \\
DiffFace & 0.02 & 0.63 & 0.65 & 0.00 & 99.75 & 0.59 & 0.00 & 1.88 & 0.02 & 0.00 & 1.22 & 0.87 & 0.11 & 100.00 \\
DreamBooth & 50.67 & 57.30 & 98.98 & 0.00 & 99.94 & 97.01 & 2.08 & 99.87 & 90.01 & 1.47 & 97.73 & 97.34 & 94.68 & 65.92 \\
FreeDoM\_I & 1.87 & 0.12 & 0.00 & 31.54 & 0.00 & 0.74 & 80.86 & 0.06 & 8.06 & 91.90 & 0.44 & 0.64 & 2.60 & 98.02 \\
LoRA & 30.50 & 30.07 & 86.26 & 0.00 & 100.00 & 82.83 & 0.22 & 98.74 & 75.62 & 0.24 & 91.37 & 95.25 & 82.24 & 90.70 \\
SDXL\_Refine & 13.66 & 21.84 & 71.42 & 44.24 & 99.37 & 90.54 & 16.66 & 99.12 & 97.08 & 21.53 & 94.37 & 98.39 & 96.35 & 78.26 \\
FreeDoM\_T & 16.19 & 0.44 & 0.00 & 93.20 & 0.06 & 0.03 & 82.32 & 0.00 & 0.77 & 88.32 & 0.06 & 0.16 & 0.36 & 99.17 \\
HPS & 90.67 & 78.95 & 97.90 & 99.09 & 100.00 & 88.19 & 0.27 & 98.99 & 67.83 & 0.32 & 94.39 & 94.39 & 76.74 & 75.53 \\
Midjourney & 83.28 & 99.78 & 99.91 & 100.00 & 100.00 & 97.96 & 42.76 & 99.87 & 96.17 & 34.35 & 98.83 & 99.82 & 98.04 & 4.79 \\
SDXL & 0.57 & 19.87 & 82.38 & 99.98 & 98.54 & 96.27 & 5.89 & 99.18 & 93.40 & 4.61 & 95.29 & 95.80 & 95.81 & 68.34 \\
\hline
\hline\thickhline
     \rowcolor{mygray}
$t=6$ & CoDiff & cycle\_diff & Imagic & DCFace & DiffFace & DreamBooth & FreeDoM\_I & LoRA & SDXL\_Refine & FreeDoM\_T & HPS & Midjourney & SDXL & real \\
\hline\hline
CoDiff & 43.03 & 3.36 & 2.63 & 0.00 & 73.48 & 1.02 & 0.00 & 2.51 & 0.00 & 0.00 & 1.52 & 0.19 & 0.82 & 99.97 \\
cycle\_diff & 21.89 & 76.55 & 72.75 & 0.00 & 100.00 & 44.31 & 0.00 & 67.84 & 7.12 & 0.00 & 56.63 & 51.51 & 18.90 & 95.87 \\
Imagic & 1.89 & 35.79 & 88.11 & 0.00 & 13.45 & 57.45 & 0.00 & 53.14 & 8.33 & 0.00 & 52.34 & 29.52 & 15.83 & 90.76 \\
DCFace & 0.00 & 0.00 & 0.00 & 100.00 & 0.00 & 0.00 & 0.00 & 0.00 & 0.00 & 0.00 & 0.00 & 0.00 & 0.00 & 100.00 \\
DiffFace & 0.00 & 2.42 & 3.44 & 0.00 & 100.00 & 3.50 & 0.00 & 6.16 & 0.25 & 0.00 & 3.57 & 6.82 & 0.78 & 99.96 \\
DreamBooth & 0.00 & 0.00 & 0.00 & 0.00 & 0.00 & 0.00 & 0.00 & 0.00 & 0.00 & 0.00 & 0.00 & 0.00 & 0.00 & 100.00 \\
FreeDoM\_I & 1.44 & 0.05 & 0.02 & 100.00 & 0.00 & 0.56 & 54.14 & 0.06 & 8.92 & 73.87 & 0.54 & 0.96 & 3.43 & 98.84 \\
LoRA & 1.54 & 5.03 & 42.83 & 0.06 & 94.80 & 41.50 & 0.00 & 83.73 & 21.01 & 0.00 & 57.47 & 63.73 & 31.72 & 99.59 \\
SDXL\_Refine & 1.09 & 31.87 & 90.97 & 47.66 & 100.00 & 93.41 & 14.71 & 99.12 & 93.33 & 11.80 & 93.84 & 96.97 & 93.97 & 58.77 \\
FreeDoM\_T & 0.00 & 0.00 & 0.00 & 0.00 & 0.00 & 0.00 & 0.00 & 0.00 & 0.00 & 0.00 & 0.00 & 0.00 & 0.00 & 100.00 \\
HPS & 81.44 & 92.35 & 99.61 & 5.57 & 100.00 & 91.46 & 0.89 & 92.71 & 44.93 & 0.32 & 91.04 & 82.95 & 60.24 & 13.29 \\
Midjourney & 0.00 & 6.37 & 60.08 & 0.00 & 93.08 & 65.60 & 0.58 & 92.15 & 71.30 & 1.35 & 69.45 & 87.43 & 72.71 & 94.99 \\
SDXL & 1.22 & 0.88 & 21.52 & 0.02 & 26.33 & 27.52 & 0.09 & 82.85 & 43.29 & 0.08 & 51.26 & 59.55 & 47.00 & 99.78 \\
\hline
\hline\thickhline
     \rowcolor{mygray}
$t=10$& CoDiff & cycle\_diff & Imagic & DCFace & DiffFace & DreamBooth & FreeDoM\_I & LoRA & SDXL\_Refine & FreeDoM\_T & HPS & Midjourney & SDXL & real \\
\hline\
CoDiff & 92.71 & 88.37 & 94.53 & 0.00 & 100.00 & 79.07 & 18.61 & 68.28 & 22.97 & 11.72 & 70.82 & 51.27 & 34.26 & 6.62 \\
cycle\_diff & 2.31 & 66.40 & 80.28 & 100.00 & 100.00 & 57.37 & 0.22 & 54.08 & 6.74 & 0.00 & 49.58 & 37.60 & 16.09 & 35.79 \\
Imagic & 0.00 & 0.00 & 0.00 & 0.00 & 0.00 & 0.00 & 0.00 & 0.00 & 0.00 & 0.00 & 0.00 & 0.00 & 0.00 & 100.00 \\
DCFace & 0.00 & 0.00 & 0.00 & 100.00 & 0.00 & 0.00 & 0.00 & 0.00 & 0.00 & 0.00 & 0.00 & 0.00 & 0.00 & 100.00 \\
DiffFace & 0.00 & 0.31 & 0.16 & 0.00 & 31.28 & 0.20 & 0.00 & 0.25 & 0.00 & 0.00 & 0.53 & 0.12 & 0.02 & 100.00 \\
DreamBooth & 2.39 & 13.55 & 54.63 & 0.00 & 32.80 & 50.01 & 0.00 & 42.46 & 7.31 & 0.12 & 38.18 & 18.13 & 13.74 & 82.48 \\
FreeDoM\_I & 100.00 & 100.00 & 100.00 & 100.00 & 100.00 & 100.00 & 100.00 & 100.00 & 100.00 & 100.00 & 100.00 & 100.00 & 100.00 & 0.00 \\
LoRA & 12.49 & 21.83 & 66.24 & 4.49 & 49.37 & 77.46 & 10.46 & 93.40 & 72.89 & 8.26 & 80.13 & 85.39 & 76.76 & 69.15 \\
SDXL\_Refine & 1.32 & 4.16 & 27.55 & 35.54 & 35.85 & 34.17 & 0.44 & 71.17 & 35.40 & 0.95 & 46.82 & 61.75 & 38.47 & 92.72 \\
FreeDoM\_T & 26.44 & 6.72 & 0.21 & 100.00 & 0.00 & 2.45 & 61.36 & 1.44 & 19.81 & 72.20 & 3.03 & 3.54 & 11.87 & 96.02 \\
HPS & 63.96 & 50.76 & 78.90 & 88.72 & 77.41 & 84.41 & 61.28 & 90.45 & 76.49 & 65.05 & 87.60 & 71.81 & 80.94 & 49.49 \\
Midjourney & 100.00 & 100.00 & 100.00 & 100.00 & 100.00 & 100.00 & 100.00 & 100.00 & 100.00 & 100.00 & 100.00 & 100.00 & 100.00 & 0.00 \\
SDXL & 20.42 & 66.31 & 97.30 & 38.89 & 100.00 & 95.81 & 17.10 & 97.61 & 84.86 & 15.29 & 95.64 & 95.15 & 87.44 & 14.73 \\
\hline
\hline\thickhline
     \rowcolor{mygray}
$t=20$& CoDiff & cycle\_diff & Imagic & DCFace & DiffFace & DreamBooth & FreeDoM\_I & LoRA & SDXL\_Refine & FreeDoM\_T & HPS & Midjourney & SDXL & real \\
\hline
CoDiff & 95.95 & 48.90 & 26.37 & 96.54 & 14.66 & 34.78 & 78.64 & 43.22 & 71.70 & 86.18 & 42.04 & 47.65 & 65.11 & 76.78 \\
cycle\_diff & 0.00 & 0.00 & 0.00 & 0.00 & 0.00 & 0.00 & 0.00 & 0.00 & 0.00 & 0.00 & 0.00 & 0.00 & 0.00 & 100.00 \\
Imagic & 14.60 & 41.45 & 80.70 & 0.00 & 66.56 & 64.71 & 22.29 & 56.47 & 19.73 & 14.10 & 55.17 & 41.99 & 27.32 & 31.07 \\
DCFace & 0.00 & 0.00 & 0.00 & 100.00 & 0.00 & 0.00 & 0.00 & 0.00 & 0.00 & 0.00 & 0.00 & 0.00 & 0.00 & 100.00 \\
DiffFace & 24.55 & 56.48 & 76.03 & 1.58 & 99.05 & 61.51 & 36.64 & 48.12 & 20.94 & 24.07 & 55.46 & 47.48 & 26.54 & 13.16 \\
DreamBooth & 60.62 & 74.61 & 94.25 & 0.00 & 83.31 & 96.09 & 67.35 & 94.41 & 87.98 & 66.72 & 93.84 & 89.80 & 89.85 & 13.71 \\
FreeDoM\_I & 2.56 & 1.18 & 0.35 & 99.91 & 0.06 & 1.02 & 1.73 & 1.88 & 5.12 & 4.17 & 1.18 & 1.58 & 4.18 & 99.82 \\
LoRA & 3.71 & 12.63 & 38.47 & 0.09 & 18.85 & 34.19 & 4.34 & 34.55 & 7.65 & 2.46 & 29.04 & 18.12 & 12.47 & 69.06 \\
SDXL\_Refine & 2.94 & 1.54 & 2.84 & 0.09 & 0.44 & 6.70 & 2.44 & 15.33 & 20.67 & 3.97 & 9.12 & 16.42 & 18.37 & 99.30 \\
FreeDoM\_T & 45.30 & 28.33 & 14.96 & 100.00 & 36.68 & 11.01 & 35.36 & 14.20 & 22.58 & 43.01 & 14.08 & 15.48 & 21.80 & 72.53 \\
HPS & 1.00 & 5.13 & 16.86 & 0.00 & 5.90 & 14.57 & 1.15 & 13.76 & 1.61 & 0.60 & 13.71 & 6.82 & 3.76 & 86.69 \\
Midjourney & 90.75 & 95.73 & 99.47 & 20.37 & 99.43 & 99.05 & 93.18 & 99.43 & 96.74 & 90.39 & 98.98 & 99.17 & 97.20 & 1.36 \\
SDXL & 100.00 & 100.00 & 100.00 & 0.00 & 100.00 & 100.00 & 100.00 & 100.00 & 100.00 & 100.00 & 100.00 & 100.00 & 100.00 & 0.00 \\
\hline
\end{tabular}
\end{adjustbox}
\end{table*}

\begin{table*}[htbp] 
\centering
\centering \setlength{\tabcolsep}{3pt}\small
\caption{Cross-model ACC on DiFF (timestep $t$ from 30 to 100).}
\label{tab:timestep3to10}
\begin{adjustbox}{max width=\textwidth}
\begin{tabular}{|l| *{14}{|r}|}
\hline\thickhline
     \rowcolor{mygray}
$t=30$ & CoDiff & cycle\_diff & Imagic & DCFace & DiffFace & DreamBooth & FreeDoM\_I & LoRA & SDXL\_Refine & FreeDoM\_T & HPS & Midjourney & SDXL & real \\
\hline
CoDiff & 56.74 & 37.03 & 26.44 & 99.96 & 30.08 & 26.14 & 42.31 & 29.59 & 39.54 & 47.82 & 29.47 & 34.16 & 38.73 & 73.99 \\
cycle\_diff & 0.02 & 1.44 & 4.39 & 0.00 & 0.70 & 2.33 & 0.44 & 1.76 & 0.04 & 0.00 & 2.71 & 0.50 & 0.38 & 93.84 \\
Imagic & 88.43 & 94.65 & 99.01 & 73.36 & 98.67 & 97.19 & 91.85 & 96.23 & 84.86 & 84.87 & 96.14 & 93.89 & 87.51 & 0.88 \\
DCFace & 0.00 & 0.00 & 0.00 & 100.00 & 0.00 & 0.00 & 0.00 & 0.00 & 0.00 & 0.00 & 0.00 & 0.00 & 0.00 & 100.00 \\
DiffFace & 0.00 & 0.11 & 0.46 & 0.00 & 0.00 & 0.13 & 0.04 & 0.13 & 0.00 & 0.00 & 0.22 & 0.05 & 0.02 & 99.70 \\
DreamBooth & 0.02 & 0.01 & 0.09 & 0.00 & 0.00 & 0.18 & 0.00 & 0.06 & 0.00 & 0.00 & 0.26 & 0.00 & 0.04 & 99.93 \\
FreeDoM\_I & 100.00 & 100.00 & 100.00 & 100.00 & 100.00 & 100.00 & 100.00 & 100.00 & 100.00 & 100.00 & 100.00 & 100.00 & 100.00 & 0.00 \\
LoRA & 3.38 & 8.18 & 19.16 & 0.00 & 7.04 & 17.38 & 6.07 & 15.26 & 5.94 & 2.26 & 16.05 & 10.11 & 7.53 & 82.95 \\
SDXL\_Refine & 0.95 & 1.85 & 2.77 & 0.00 & 2.22 & 3.14 & 1.24 & 2.32 & 1.46 & 1.07 & 2.77 & 2.37 & 1.65 & 97.21 \\
FreeDoM\_T & 98.41 & 95.46 & 91.53 & 100.00 & 93.46 & 91.57 & 97.25 & 93.97 & 97.65 & 98.49 & 93.20 & 96.39 & 97.29 & 8.69 \\
HPS & 0.00 & 0.04 & 0.44 & 0.00 & 0.06 & 0.38 & 0.00 & 0.31 & 0.02 & 0.00 & 0.62 & 0.02 & 0.02 & 99.79 \\
Midjourney & 7.94 & 18.25 & 36.18 & 96.65 & 25.44 & 30.97 & 12.49 & 26.26 & 12.09 & 7.15 & 27.70 & 24.69 & 15.05 & 61.84 \\
SDXL & 100.00 & 99.93 & 99.98 & 0.00 & 100.00 & 100.00 & 99.96 & 100.00 & 99.95 & 99.72 & 99.94 & 99.95 & 99.96 & 0.01 \\
\hline
\hline\thickhline
     \rowcolor{mygray}
$t=40$ & CoDiff & cycle\_diff & Imagic & DCFace & DiffFace & DreamBooth & FreeDoM\_I & LoRA & SDXL\_Refine & FreeDoM\_T & HPS & Midjourney & SDXL & real \\
\hline
CoDiff & 82.16 & 76.16 & 75.85 & 1.88 & 77.79 & 74.34 & 78.33 & 73.81 & 74.37 & 79.11 & 73.93 & 74.25 & 75.25 & 21.93 \\
cycle\_diff & 5.00 & 15.95 & 31.36 & 0.00 & 15.74 & 24.35 & 9.66 & 17.15 & 4.96 & 3.42 & 18.31 & 10.55 & 7.06 & 64.41 \\
Imagic & 12.14 & 23.94 & 41.98 & 34.31 & 22.97 & 35.98 & 17.46 & 26.32 & 12.54 & 10.25 & 28.57 & 19.74 & 15.18 & 56.78 \\
DCFace & 0.00 & 0.00 & 0.00 & 100.00 & 0.00 & 0.00 & 0.00 & 0.00 & 0.00 & 0.00 & 0.00 & 0.00 & 0.00 & 100.00 \\
DiffFace & 5.90 & 17.23 & 34.31 & 0.00 & 24.30 & 24.87 & 11.08 & 16.90 & 5.96 & 4.69 & 20.14 & 12.96 & 8.50 & 60.27 \\
DreamBooth & 6.92 & 19.31 & 42.55 & 30.11 & 18.91 & 37.62 & 12.80 & 25.31 & 8.76 & 5.88 & 27.90 & 16.96 & 12.49 & 56.87 \\
FreeDoM\_I & 91.17 & 90.38 & 90.53 & 7.24 & 91.05 & 89.16 & 90.25 & 90.89 & 91.67 & 91.46 & 90.38 & 90.07 & 91.59 & 10.23 \\
LoRA & 0.00 & 0.00 & 0.00 & 0.00 & 0.00 & 0.00 & 0.00 & 0.00 & 0.00 & 0.00 & 0.00 & 0.00 & 0.00 & 100.00 \\
SDXL\_Refine & 99.88 & 100.00 & 99.95 & 3.85 & 100.00 & 99.95 & 99.96 & 99.87 & 99.88 & 99.88 & 99.96 & 99.93 & 99.93 & 0.06 \\
FreeDoM\_T & 8.46 & 9.33 & 10.69 & 99.78 & 8.12 & 9.99 & 10.15 & 10.36 & 8.31 & 9.61 & 9.57 & 8.50 & 8.73 & 89.67 \\
HPS & 2.91 & 10.38 & 24.17 & 0.28 & 9.39 & 21.70 & 6.65 & 15.20 & 4.30 & 2.98 & 16.38 & 8.84 & 6.37 & 74.86 \\
Midjourney & 0.00 & 0.00 & 0.00 & 0.00 & 0.00 & 0.00 & 0.00 & 0.00 & 0.00 & 0.00 & 0.00 & 0.00 & 0.00 & 100.00 \\
SDXL & 85.97 & 85.82 & 87.62 & 0.13 & 86.17 & 87.07 & 87.24 & 87.69 & 87.71 & 86.38 & 87.40 & 86.61 & 87.36 & 12.30 \\
\hline
\hline\thickhline
     \rowcolor{mygray}
$t=50$& CoDiff & cycle\_diff & Imagic & DCFace & DiffFace & DreamBooth & FreeDoM\_I & LoRA & SDXL\_Refine & FreeDoM\_T & HPS & Midjourney & SDXL & real \\
\hline\hline
CoDiff & 4.10 & 3.29 & 2.82 & 0.71 & 3.43 & 3.25 & 3.32 & 2.89 & 3.19 & 4.13 & 2.95 & 3.29 & 3.54 & 97.34 \\
cycle\_diff & 45.90 & 54.48 & 61.14 & 17.95 & 55.27 & 58.78 & 48.78 & 52.01 & 43.89 & 44.56 & 53.91 & 50.26 & 45.39 & 37.93 \\
Imagic & 99.90 & 99.88 & 100.00 & 17.48 & 100.00 & 99.97 & 99.82 & 99.94 & 99.79 & 99.76 & 99.94 & 99.91 & 99.82 & 0.09 \\
DCFace & 0.00 & 0.00 & 0.00 & 100.00 & 0.00 & 0.00 & 0.00 & 0.00 & 0.00 & 0.00 & 0.00 & 0.00 & 0.00 & 100.00 \\
DiffFace & 99.88 & 99.96 & 99.93 & 98.60 & 99.87 & 100.00 & 99.91 & 99.94 & 99.88 & 99.92 & 99.93 & 99.93 & 99.82 & 0.04 \\
DreamBooth & 78.61 & 84.72 & 90.19 & 0.15 & 83.31 & 88.47 & 81.26 & 85.24 & 78.78 & 78.04 & 87.04 & 83.35 & 80.08 & 10.25 \\
FreeDoM\_I & 0.77 & 0.99 & 0.85 & 70.49 & 0.63 & 1.02 & 0.75 & 0.94 & 0.78 & 0.60 & 0.80 & 0.71 & 0.62 & 99.10 \\
LoRA & 1.97 & 3.25 & 4.99 & 0.00 & 2.16 & 4.37 & 2.79 & 3.77 & 2.35 & 2.10 & 4.12 & 3.13 & 2.34 & 95.23 \\
SDXL\_Refine & 7.34 & 7.29 & 6.93 & 0.00 & 5.39 & 6.49 & 7.40 & 7.16 & 9.18 & 8.70 & 6.62 & 8.64 & 9.20 & 93.26 \\
FreeDoM\_T & 66.64 & 66.92 & 67.88 & 100.00 & 65.99 & 68.13 & 67.74 & 68.03 & 66.83 & 68.07 & 67.92 & 67.52 & 66.79 & 31.38 \\
HPS & 3.33 & 7.14 & 12.75 & 0.00 & 6.54 & 12.16 & 5.27 & 8.79 & 4.07 & 3.02 & 10.18 & 5.83 & 4.34 & 87.39 \\
Midjourney & 10.07 & 15.95 & 21.59 & 0.19 & 15.48 & 19.42 & 14.71 & 17.27 & 11.68 & 10.41 & 17.27 & 14.36 & 12.49 & 78.45 \\
SDXL & 8.71 & 9.92 & 10.23 & 31.50 & 9.33 & 10.84 & 9.48 & 9.80 & 9.22 & 9.57 & 10.32 & 10.57 & 9.53 & 89.22 \\
\hline
\hline\thickhline
     \rowcolor{mygray}
$t=100$ & CoDiff & cycle\_diff & Imagic & DCFace & DiffFace & DreamBooth & FreeDoM\_I & LoRA & SDXL\_Refine & FreeDoM\_T & HPS & Midjourney & SDXL & real \\
\hline\hline
CoDiff & 89.53 & 89.47 & 89.15 & 100.00 & 90.10 & 89.45 & 88.39 & 89.64 & 88.48 & 88.64 & 88.35 & 88.72 & 89.63 & 11.00 \\
cycle\_diff & 99.40 & 99.41 & 99.52 & 37.42 & 99.68 & 99.46 & 99.65 & 99.37 & 99.55 & 99.52 & 99.58 & 99.36 & 99.67 & 0.52 \\
Imagic & 9.58 & 9.46 & 10.60 & 0.00 & 9.07 & 10.38 & 8.77 & 9.99 & 9.01 & 7.43 & 8.95 & 9.19 & 8.13 & 90.00 \\
DCFace & 0.00 & 0.00 & 0.00 & 100.00 & 0.00 & 0.00 & 0.00 & 0.00 & 0.00 & 0.00 & 0.00 & 0.00 & 0.00 & 100.00 \\
DiffFace & 2.79 & 2.68 & 2.96 & 64.51 & 2.35 & 3.02 & 3.37 & 3.08 & 2.82 & 2.86 & 2.75 & 2.71 & 2.60 & 97.00 \\
DreamBooth & 71.49 & 72.13 & 73.40 & 0.00 & 71.32 & 73.91 & 71.95 & 72.99 & 70.77 & 71.41 & 71.87 & 70.94 & 72.08 & 27.47 \\
FreeDoM\_I & 22.86 & 22.89 & 22.95 & 0.00 & 22.91 & 23.00 & 21.53 & 23.24 & 22.49 & 21.09 & 22.79 & 22.33 & 23.55 & 76.84 \\
LoRA & 1.49 & 1.44 & 1.55 & 0.00 & 1.02 & 1.51 & 1.42 & 2.26 & 1.30 & 0.95 & 1.38 & 1.52 & 1.34 & 98.72 \\
SDXL\_Refine & 16.64 & 16.33 & 16.92 & 52.04 & 18.97 & 17.38 & 17.46 & 16.58 & 16.44 & 15.97 & 16.09 & 17.28 & 16.21 & 83.51 \\
FreeDoM\_T & 1.67 & 1.66 & 1.20 & 0.00 & 2.35 & 1.87 & 1.55 & 2.20 & 1.66 & 1.43 & 1.44 & 1.61 & 1.49 & 98.16 \\
HPS & 100.00 & 100.00 & 100.00 & 100.00 & 100.00 & 100.00 & 100.00 & 100.00 & 100.00 & 100.00 & 100.00 & 100.00 & 100.00 & 0.00 \\
Midjourney & 17.09 & 16.72 & 16.79 & 0.00 & 17.13 & 17.12 & 16.08 & 14.45 & 16.23 & 16.60 & 16.44 & 15.69 & 15.56 & 82.86 \\
SDXL & 14.00 & 14.94 & 14.45 & 100.00 & 15.80 & 15.26 & 15.64 & 14.45 & 14.36 & 13.94 & 14.61 & 13.90 & 15.00 & 85.19 \\
\hline

\end{tabular}
\end{adjustbox}
\end{table*}
\begin{table*}[htbp] 
\centering
\centering \setlength{\tabcolsep}{3pt}\small
\caption{Cross-model ACC on DiFF (timestep $t$ from 200 to 800).}
\label{tab:timestep20to80}
\begin{adjustbox}{max width=\textwidth}
\begin{tabular}{|l| *{14}{|r}|}
\hline\thickhline
     \rowcolor{mygray}
$t=200$ & CoDiff & cycle\_diff & Imagic & DCFace & DiffFace & DreamBooth & FreeDoM\_I & LoRA & SDXL\_Refine & FreeDoM\_T & HPS & Midjourney & SDXL & real \\
\hline\hline
CoDiff & 8.48 & 8.46 & 8.50 & 0.02 & 10.15 & 7.92 & 9.66 & 8.61 & 8.74 & 8.82 & 9.25 & 7.95 & 9.28 & 91.38 \\
cycle\_diff & 35.00 & 34.68 & 34.17 & 0.00 & 34.52 & 35.11 & 34.03 & 35.18 & 34.62 & 34.15 & 34.62 & 34.27 & 33.35 & 65.38 \\
Imagic & 96.69 & 96.24 & 96.49 & 99.81 & 96.45 & 96.22 & 96.85 & 96.55 & 96.77 & 96.39 & 96.60 & 96.44 & 96.82 & 3.45 \\
DCFace & 0.00 & 0.00 & 0.00 & 100.00 & 0.00 & 0.00 & 0.00 & 0.00 & 0.00 & 0.00 & 0.00 & 0.00 & 0.00 & 100.00 \\
DiffFace & 85.20 & 85.13 & 85.18 & 97.34 & 85.47 & 84.85 & 84.49 & 83.29 & 86.04 & 84.23 & 86.32 & 84.61 & 84.66 & 14.53 \\
DreamBooth & 93.93 & 93.70 & 94.00 & 84.60 & 94.42 & 94.22 & 94.37 & 94.66 & 94.40 & 94.24 & 93.74 & 93.55 & 93.77 & 6.18 \\
FreeDoM\_I & 50.17 & 49.98 & 49.39 & 19.33 & 49.49 & 51.01 & 50.78 & 50.50 & 48.94 & 49.32 & 50.11 & 50.27 & 49.58 & 50.15 \\
LoRA & 1.02 & 1.44 & 1.62 & 0.00 & 1.02 & 1.33 & 1.46 & 1.26 & 1.52 & 1.47 & 1.36 & 1.63 & 1.40 & 98.55 \\
SDXL\_Refine & 51.99 & 51.33 & 51.84 & 59.32 & 51.97 & 50.88 & 51.04 & 50.31 & 51.19 & 51.11 & 51.36 & 53.34 & 51.67 & 47.19 \\
FreeDoM\_T & 21.59 & 23.49 & 22.70 & 53.96 & 21.32 & 22.67 & 23.13 & 23.05 & 22.53 & 23.75 & 22.37 & 22.54 & 21.79 & 76.52 \\
HPS & 39.10 & 37.62 & 37.34 & 0.89 & 38.39 & 38.28 & 37.35 & 37.06 & 39.04 & 37.85 & 37.94 & 37.21 & 38.49 & 61.75 \\
Midjourney & 12.24 & 13.32 & 13.09 & 0.00 & 13.83 & 13.03 & 14.09 & 11.81 & 13.88 & 12.43 & 13.39 & 14.01 & 14.11 & 86.27 \\
SDXL & 81.12 & 80.73 & 79.91 & 39.77 & 82.99 & 79.99 & 80.99 & 80.72 & 80.65 & 79.11 & 80.54 & 80.24 & 80.25 & 19.43 \\
\hline
\hline\thickhline
     \rowcolor{mygray}
$t=400$& CoDiff & cycle\_diff & Imagic & DCFace & DiffFace & DreamBooth & FreeDoM\_I & LoRA & SDXL\_Refine & FreeDoM\_T & HPS & Midjourney & SDXL & real \\
\hline\hline
CoDiff & 93.28 & 93.14 & 94.00 & 89.52 & 94.92 & 93.28 & 92.82 & 92.53 & 93.76 & 93.29 & 93.68 & 93.64 & 93.48 & 6.57 \\
cycle\_diff & 65.05 & 64.89 & 64.33 & 83.19 & 64.85 & 63.99 & 64.64 & 64.13 & 65.92 & 62.99 & 64.04 & 64.72 & 65.45 & 35.21 \\
Imagic & 71.72 & 71.02 & 71.32 & 1.66 & 71.32 & 70.53 & 69.74 & 69.66 & 71.11 & 70.14 & 71.61 & 71.99 & 71.64 & 28.74 \\
DCFace & 0.00 & 0.00 & 0.00 & 100.00 & 0.00 & 0.00 & 0.00 & 0.00 & 0.00 & 0.00 & 0.00 & 0.00 & 0.00 & 100.00 \\
DiffFace & 48.91 & 48.44 & 46.43 & 94.32 & 45.69 & 48.02 & 47.36 & 46.36 & 48.15 & 48.57 & 46.78 & 48.27 & 47.77 & 52.31 \\
DreamBooth & 1.02 & 0.72 & 0.72 & 2.61 & 0.76 & 0.49 & 0.80 & 0.88 & 0.84 & 0.68 & 0.86 & 0.92 & 0.69 & 99.12 \\
FreeDoM\_I & 56.89 & 57.53 & 57.05 & 27.03 & 56.98 & 56.04 & 56.18 & 57.85 & 56.36 & 58.18 & 57.30 & 56.56 & 56.45 & 44.42 \\
LoRA & 4.10 & 4.23 & 4.09 & 0.37 & 4.00 & 4.09 & 3.54 & 4.08 & 4.12 & 3.65 & 4.18 & 4.37 & 4.18 & 95.69 \\
SDXL\_Refine & 65.15 & 65.19 & 65.32 & 99.87 & 64.78 & 64.22 & 64.73 & 64.95 & 66.68 & 65.17 & 66.16 & 66.16 & 65.70 & 34.29 \\
FreeDoM\_T & 58.83 & 59.10 & 60.12 & 82.70 & 57.30 & 60.64 & 61.05 & 57.91 & 59.75 & 60.01 & 59.04 & 59.36 & 59.39 & 40.61 \\
HPS & 40.85 & 40.13 & 41.75 & 76.63 & 41.37 & 40.74 & 42.05 & 41.58 & 41.88 & 41.74 & 40.88 & 39.99 & 40.40 & 59.94 \\
Midjourney & 94.98 & 95.47 & 95.91 & 18.82 & 95.37 & 95.48 & 94.82 & 96.11 & 95.61 & 95.04 & 95.32 & 95.17 & 95.12 & 4.82 \\
SDXL & 58.93 & 60.01 & 59.46 & 72.93 & 59.01 & 59.95 & 59.81 & 60.30 & 59.62 & 59.97 & 61.12 & 61.40 & 60.66 & 39.62 \\
\hline
\hline\thickhline
     \rowcolor{mygray}
$t=600$& CoDiff & cycle\_diff & Imagic & DCFace & DiffFace & DreamBooth & FreeDoM\_I & LoRA & SDXL\_Refine & FreeDoM\_T & HPS & Midjourney & SDXL & real \\
\hline\hline
CoDiff & 97.06 & 97.50 & 97.37 & 96.22 & 97.08 & 97.37 & 97.08 & 97.17 & 96.93 & 96.54 & 96.77 & 96.95 & 97.06 & 2.78 \\
cycle\_diff & 74.98 & 75.29 & 75.53 & 91.86 & 72.97 & 74.85 & 75.94 & 74.81 & 75.66 & 75.14 & 75.05 & 74.96 & 74.49 & 24.70 \\
Imagic & 29.33 & 29.09 & 29.28 & 68.93 & 29.57 & 29.80 & 29.91 & 29.08 & 28.61 & 30.22 & 28.97 & 30.14 & 29.10 & 71.12 \\
DCFace & 0.00 & 0.00 & 0.00 & 100.00 & 0.00 & 0.00 & 0.00 & 0.00 & 0.00 & 0.00 & 0.00 & 0.00 & 0.00 & 100.00 \\
DiffFace & 47.39 & 48.86 & 47.56 & 34.00 & 48.03 & 49.09 & 47.41 & 48.12 & 47.60 & 47.10 & 46.81 & 48.50 & 48.35 & 52.21 \\
DreamBooth & 58.86 & 59.44 & 58.67 & 81.23 & 57.68 & 59.88 & 58.84 & 58.23 & 58.50 & 57.31 & 59.37 & 58.46 & 58.02 & 41.64 \\
FreeDoM\_I & 7.49 & 7.51 & 7.67 & 19.96 & 8.19 & 8.02 & 8.37 & 9.80 & 7.17 & 7.86 & 7.89 & 7.70 & 8.04 & 91.68 \\
LoRA & 36.72 & 38.60 & 36.74 & 46.77 & 37.69 & 37.72 & 36.60 & 40.39 & 37.63 & 36.42 & 36.62 & 36.21 & 37.20 & 62.40 \\
SDXL\_Refine & 78.93 & 78.96 & 78.09 & 90.26 & 77.54 & 78.87 & 79.13 & 80.03 & 78.10 & 77.20 & 78.28 & 78.31 & 78.20 & 22.35 \\
FreeDoM\_T & 27.31 & 27.45 & 26.99 & 87.38 & 26.90 & 25.76 & 27.29 & 28.14 & 25.34 & 27.08 & 26.29 & 26.49 & 25.29 & 73.50 \\
HPS & 79.83 & 81.26 & 81.30 & 74.14 & 80.01 & 81.37 & 81.57 & 80.72 & 79.97 & 81.29 & 81.04 & 81.39 & 80.79 & 18.84 \\
Midjourney & 82.24 & 83.07 & 83.12 & 71.53 & 81.79 & 82.72 & 82.28 & 81.97 & 82.77 & 82.80 & 83.21 & 82.04 & 82.01 & 18.03 \\
SDXL & 59.48 & 60.14 & 61.74 & 75.48 & 60.53 & 60.18 & 60.43 & 59.74 & 59.35 & 60.72 & 59.70 & 60.46 & 58.77 & 40.42 \\
\hline
\hline\thickhline
     \rowcolor{mygray}
$t=800$& CoDiff & cycle\_diff & Imagic & DCFace & DiffFace & DreamBooth & FreeDoM\_I & LoRA & SDXL\_Refine & FreeDoM\_T & HPS & Midjourney & SDXL & real \\
\hline\hline
CoDiff & 96.22 & 95.65 & 95.73 & 88.40 & 95.49 & 96.32 & 95.75 & 96.48 & 96.83 & 95.75 & 95.81 & 95.87 & 96.30 & 4.15 \\
cycle\_diff & 77.31 & 76.98 & 77.63 & 49.41 & 76.71 & 77.33 & 77.45 & 79.27 & 76.55 & 76.65 & 76.70 & 77.19 & 76.54 & 22.63 \\
Imagic & 33.18 & 33.54 & 33.53 & 22.58 & 31.35 & 33.07 & 32.43 & 31.09 & 33.08 & 33.80 & 32.46 & 32.85 & 33.59 & 67.34 \\
DCFace & 0.00 & 0.00 & 0.00 & 100.00 & 0.00 & 0.00 & 0.00 & 0.00 & 0.00 & 0.00 & 0.00 & 0.00 & 0.00 & 100.00 \\
DiffFace & 92.36 & 92.92 & 92.75 & 97.45 & 92.58 & 92.97 & 91.89 & 92.21 & 91.85 & 92.10 & 92.01 & 92.35 & 92.08 & 7.78 \\
DreamBooth & 58.31 & 57.98 & 58.00 & 3.61 & 57.55 & 59.26 & 57.60 & 57.91 & 57.73 & 57.35 & 57.00 & 56.49 & 58.90 & 42.35 \\
FreeDoM\_I & 27.19 & 25.92 & 26.41 & 28.73 & 27.86 & 27.37 & 28.27 & 25.82 & 24.68 & 27.60 & 25.78 & 26.10 & 26.71 & 73.79 \\
LoRA & 48.03 & 48.46 & 48.07 & 40.05 & 47.91 & 46.61 & 49.05 & 48.56 & 47.37 & 48.65 & 47.69 & 47.85 & 48.26 & 52.17 \\
SDXL\_Refine & 76.92 & 77.72 & 75.71 & 48.95 & 78.36 & 77.10 & 76.78 & 77.20 & 77.94 & 77.44 & 78.32 & 78.17 & 78.43 & 22.68 \\
FreeDoM\_T & 53.81 & 53.98 & 56.06 & 54.55 & 55.20 & 56.10 & 55.16 & 54.77 & 54.86 & 54.29 & 54.33 & 55.78 & 54.05 & 45.62 \\
HPS & 40.52 & 40.56 & 39.55 & 63.69 & 39.21 & 40.76 & 40.01 & 40.01 & 40.25 & 40.98 & 40.35 & 41.65 & 40.03 & 59.57 \\
Midjourney & 44.80 & 45.48 & 43.15 & 72.91 & 45.43 & 43.70 & 45.99 & 42.53 & 44.12 & 44.80 & 44.87 & 44.77 & 44.79 & 55.74 \\
SDXL & 64.30 & 64.80 & 65.14 & 41.20 & 65.42 & 65.42 & 66.02 & 64.70 & 63.76 & 63.26 & 65.24 & 64.78 & 64.25 & 35.12 \\
\hline
\end{tabular}
\end{adjustbox}
\end{table*}
\begin{table*}[htbp] 
\centering
\centering \setlength{\tabcolsep}{3pt}\small
\caption{Cross-model ACC on DiFF (timestep $t$ at 1000).}
\label{tab:timestep100}
\begin{adjustbox}{max width=\textwidth}
\begin{tabular}{|l| *{14}{|r}|}
\hline\thickhline
     \rowcolor{mygray}
$t=1000$& CoDiff & cycle\_diff & Imagic & DCFace & DiffFace & DreamBooth & FreeDoM\_I & LoRA & SDXL\_Refine & FreeDoM\_T & HPS & Midjourney & SDXL & real \\
\hline\hline
CoDiff & 96.22 & 95.65 & 95.73 & 88.40 & 95.49 & 96.32 & 95.75 & 96.48 & 96.83 & 95.75 & 95.81 & 95.87 & 96.30 & 4.15 \\
cycle\_diff & 77.31 & 76.98 & 77.63 & 49.41 & 76.71 & 77.33 & 77.45 & 79.27 & 76.55 & 76.65 & 76.70 & 77.19 & 76.54 & 22.63 \\
Imagic & 33.18 & 33.54 & 33.53 & 22.58 & 31.35 & 33.07 & 32.43 & 31.09 & 33.08 & 33.80 & 32.46 & 32.85 & 33.59 & 67.34 \\
DCFace & 0.00 & 0.00 & 0.00 & 100.00 & 0.00 & 0.00 & 0.00 & 0.00 & 0.00 & 0.00 & 0.00 & 0.00 & 0.00 & 100.00 \\
DiffFace & 92.36 & 92.92 & 92.75 & 97.45 & 92.58 & 92.97 & 91.89 & 92.21 & 91.85 & 92.10 & 92.01 & 92.35 & 92.08 & 7.78 \\
DreamBooth & 58.31 & 57.98 & 58.00 & 3.61 & 57.55 & 59.26 & 57.60 & 57.91 & 57.73 & 57.35 & 57.00 & 56.49 & 58.90 & 42.35 \\
FreeDoM\_I & 27.19 & 25.92 & 26.41 & 28.73 & 27.86 & 27.37 & 28.27 & 25.82 & 24.68 & 27.60 & 25.78 & 26.10 & 26.71 & 73.79 \\
LoRA & 48.03 & 48.46 & 48.07 & 40.05 & 47.91 & 46.61 & 49.05 & 48.56 & 47.37 & 48.65 & 47.69 & 47.85 & 48.26 & 52.17 \\
SDXL\_Refine & 76.92 & 77.72 & 75.71 & 48.95 & 78.36 & 77.10 & 76.78 & 77.20 & 77.94 & 77.44 & 78.32 & 78.17 & 78.43 & 22.68 \\
FreeDoM\_T & 53.81 & 53.98 & 56.06 & 54.55 & 55.20 & 56.10 & 55.16 & 54.77 & 54.86 & 54.29 & 54.33 & 55.78 & 54.05 & 45.62 \\
HPS & 40.52 & 40.56 & 39.55 & 63.69 & 39.21 & 40.76 & 40.01 & 40.01 & 40.25 & 40.98 & 40.35 & 41.65 & 40.03 & 59.57 \\
Midjourney & 44.80 & 45.48 & 43.15 & 72.91 & 45.43 & 43.70 & 45.99 & 42.53 & 44.12 & 44.80 & 44.87 & 44.77 & 44.79 & 55.74 \\
SDXL & 64.30 & 64.80 & 65.14 & 41.20 & 65.42 & 65.42 & 66.02 & 64.70 & 63.76 & 63.26 & 65.24 & 64.78 & 64.25 & 35.12 \\
\hline
\end{tabular}
\end{adjustbox}
\end{table*}
\end{document}